\documentclass{article}

\PassOptionsToPackage{numbers, compress}{natbib}

\usepackage[final]{neurips_2024}

\usepackage[utf8]{inputenc} 
\usepackage[T1]{fontenc}    
\usepackage{hyperref}       
\usepackage{url}            
\usepackage{booktabs}       
\usepackage{amsfonts}       
\usepackage{nicefrac}       
\usepackage{microtype}      
\usepackage{xcolor}         
\usepackage{graphicx, wrapfig}
\usepackage{subcaption}
\newcommand{\ie}{\emph{i.e.}}
\usepackage{amsmath}
\usepackage{amsthm}
\usepackage{booktabs}
\usepackage{algorithm}
\usepackage{algorithmic}
\usepackage{subcaption}
\usepackage{bbm}
\usepackage{amssymb}
\usepackage{dsfont}
\usepackage{colortbl}
\newcommand{\darkgreen}[1]{\textcolor[rgb]{0.00,0.90,0.00}{#1}}

\hypersetup{
    colorlinks=true,
    linkcolor=red,
    filecolor=magenta,      
    urlcolor=magenta,
}

\title{GoMatching: A Simple Baseline for Video Text Spotting via Long and Short Term Matching}

\author{%
  Haibin He$^{1*}$, Maoyuan Ye$^{1}$\thanks{Equal contribution, $\dagger$ Corresponding author.} , Jing Zhang$^{1\dagger}$, Juhua Liu$^{1\dagger}$, Bo Du$^{1}$, Dacheng Tao$^{2}$ \\
  \selectfont $^{1}$ School of Computer Science, National Engineering Research Center for Multimedia Software, \\ 
  \selectfont and Institute of Artificial Intelligence, Wuhan University, China \\
  \selectfont $^{2}$ College of Computing \& Data Science at Nanyang Technological University \\
  \texttt{\{haibinhe, yemaoyuan, liujuhua, dubo\}@whu.edu.com} \\
  \texttt{jingzhang.cv@gmail.com, dacheng.tao@gmail.com} \\
  \texttt{\href{https://github.com/Hxyz-123/GoMatching}{https://github.com/Hxyz-123/GoMatching}} \\
}

\begin{document}

\maketitle

\begin{abstract}
Beyond the text detection and recognition tasks in image text spotting, video text spotting presents an augmented challenge with the inclusion of tracking. While advanced end-to-end trainable methods have shown commendable performance, the pursuit of multi-task optimization may pose the risk of producing sub-optimal outcomes for individual tasks.
In this paper, we identify a main bottleneck in the state-of-the-art video text spotter: the limited recognition capability. 
In response to this issue, we propose to efficiently turn an off-the-shelf query-based image text spotter into a specialist on video and present a simple baseline termed GoMatching, which focuses the training efforts on tracking while maintaining strong recognition performance.
To adapt the image text spotter to video datasets, we add a rescoring head to rescore each detected instance's confidence via efficient tuning, leading to a better tracking candidate pool. 
Additionally, we design a long-short term matching module, termed LST-Matcher, to enhance the spotter's tracking capability by integrating both long- and short-term matching results via Transformer.
Based on the above simple designs, GoMatching delivers new records on ICDAR15-video, DSText, BOVText, and our proposed novel test with arbitrary-shaped text termed ArTVideo, which demonstrates GoMatching's capability to accommodate general, dense, small, arbitrary-shaped, Chinese and English text scenarios while saving considerable training budgets.
\end{abstract}

\section{Introduction}
Text spotting has received increasing attention due to its various applications, such as video retrieval~\cite{Dong2021Dual} and autonomous driving~\cite{Zhang2021autodrive}. 
Recently, numerous image text spotting (ITS) 
methods~\cite{liu2023spts,zhang2022text,ye2023deepsolo,huang2023estextspotter}
that simultaneously tackle text detection and recognition, have attained extraordinary accomplishment. 
In the video realm, video text spotting (VTS) involves a tracking task additionally. Although VTS methods~\cite{wang2017end,cheng2019you,cheng2020free,Wu2021BOVText,Wu2022CoText,Wu2022TransDETR} make significant progress, a substantial discrepancy persists when compared to ITS. 
We observe that the text recognition proficiency of VTS models is far inferior to ITS models.
To investigate this issue, we compare the state-of-the-art (SOTA) VTS model TransDETR~\cite{Wu2022TransDETR} and ITS model Deepsolo~\cite{ye2023deepsolo} for image-level text spotting performance on ICDAR15-video~\cite{Karatzas2015ICDAR15} and our established ArTVideo (\textit{i.e.}, \textbf{Ar}bitrary-shaped \textbf{T}ext in \textbf{Video}) test set (Sec.\ref{exp: 'dataset'}) which comprises about 30\% curved text. 

\begin{figure*}[t]
\centering
  \subfloat[]{\includegraphics[width=0.48\textwidth]{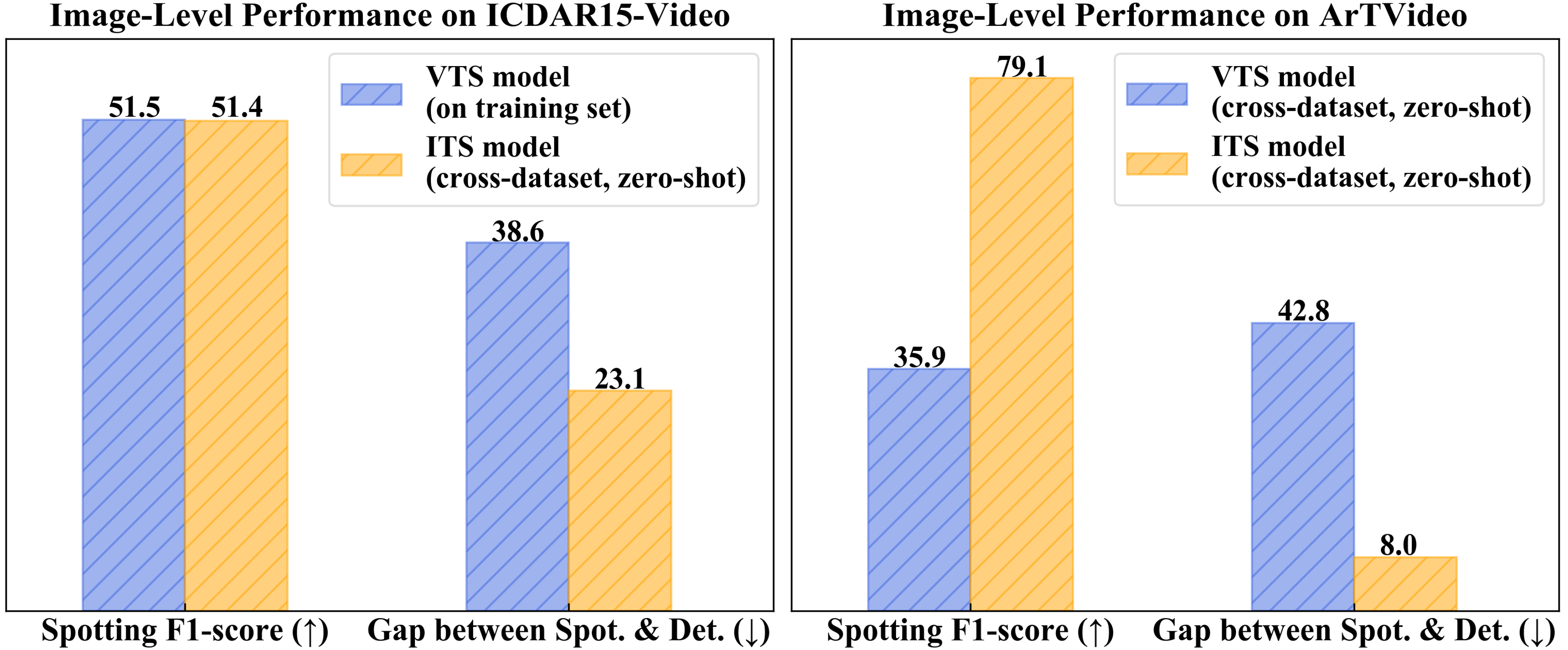}}
  \hfill
  \subfloat[]{\includegraphics[width=0.47\textwidth]{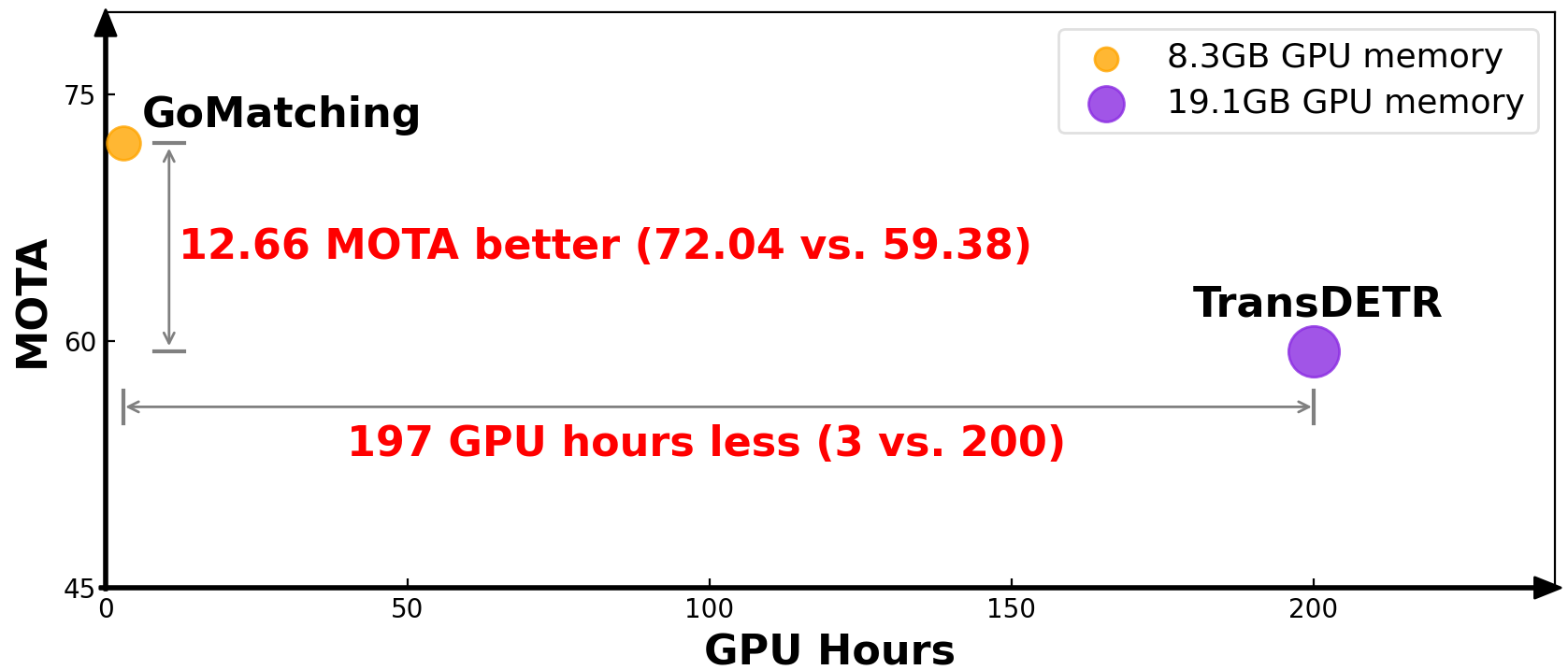}}
  \caption{(a) `Gap between Spot. \& Det.': the gap between spotting and detection F1-score. As the spotting task involves recognizing the results of the detection process, the detection score is indeed the upper bound of spotting performance. The larger the gap, the poorer the recognition ability. Compared to the ITS model (Deepsolo~\cite{ye2023deepsolo}), the VTS model (TransDETR~\cite{Wu2022TransDETR}) presents unsatisfactory image-level text spotting F1-scores, which lag far behind its detection performance, especially on ArTVideo with curved text. It indicates recognition capability is a main bottleneck in the VTS model.
  (b) GoMatching outperforms TransDETR by over 12 MOTA on ICDAR15-video while saving 197 training GPU hours and 10.8GB memory. Notice that since the pre-training strategies and settings vary between TransDETR and GoMatching, the comparison is focused on the fine-tuning stage.}
 \label{fig:1}
 \vspace{-8pt}
\end{figure*}

As illustrated in Fig.~\ref{fig:1}(a), even when evaluating the image-level spotting performance on the VTS model's training set, the F1-score of TransDETR is only comparable to the zero-shot performance of Deepsolo. 
The performance of the VTS model on ArTVideo is much worse. Moreover, there is a huge gap between the spotting and detection-only performance of the VTS model, which indicates that the recognition capability is the main bottleneck.
We attribute this discrepancy to two key aspects: 1) the model architecture and 2) the training data. 
First, in terms of model architecture, ITS studies~\cite{ye2023deepsolo,huang2023estextspotter} have presented the advantages of employing advanced query formulation for text spotting in DETR frameworks~\cite{carion2020end,zhu2020deformable}.
In contrast, existing Transformer-based VTS models still rely on Region of Interest (RoI) components or simply cropping detected text regions for recognition. 
On the other hand, some studies \cite{zhang2023motrv2,yu2023motrv3} have indicated that there exists optimization conflict in detection and association during the end-to-end training of MOTR~\cite{zeng2022motr}. We hold that TransDETR~\cite{Wu2022TransDETR}, which further incorporates text recognition into MOTR-based architecture, may also suffer from optimization conflict.
Second, regarding the training data, most text instances in current video datasets~\cite{Karatzas2015ICDAR15,Wu2021BOVText,Wu2023DSText} are straight or oriented, and the bounding box labels are only quadrilateral, which constrains the data diversity and recognition performance as well. 
Overall, the limitations in model architecture and data probably lead to the unsatisfactory text spotting performance of the SOTA VTS model. 
Hence, \textbf{\textit{leveraging model and data knowledge from ITS presents considerable value for VTS}}.
 
To achieve this, a straightforward approach is to take an off-the-shelf SOTA image text spotter and focus the training efforts on tracking across frames, akin to tracking-by-detection methods. 
An important question is how to efficiently incorporate a RoI-free image text spotter for VTS.  
In this paper, we propose a simple baseline via lon\textbf{G} and sh\textbf{o}rt term \textbf{Matching}, termed \textbf{GoMatching}, which leverages an off-the-shelf RoI-free image text spotter to identify text from each single frame and associates text instances across frames with a strong tracker. 

Specifically, we select the state-of-the-art DeepSolo~\cite{ye2023deepsolo} as the image text spotter and design a \textbf{L}ong-\textbf{S}hort \textbf{T}erm Matching-based tracker termed \textbf{LST-Matcher}.
Initially, to adapt the DeepSolo to video datasets while preserving its inherent knowledge, we freeze Deepsolo and introduce a rescoring mechanism. This mechanism entails training an additional lightweight text classifier called rescoring head via efficient tuning, and recalibrating confidence scores for detected instances to mitigate performance degradation caused by the image-video domain gap. The final score for each instance is determined by a fusion operation between the original score provided by the image text spotter and the calibrated score acquired from the rescoring head. 
The identified text instances are then sent to LST-Matcher for association. 
LST-Matcher can effectively harnesses both long- and short-term information, making it a highly capable tracker.
As a result, our baseline significantly surpasses existing SOTA methods by a large margin with much lower training costs, as shown in Fig.~\ref{fig:1}(b). 

In summary, the contribution of this paper is threefold. \textbf{1)} We identify the limitations in current VTS methods and propose a novel and simple baseline, which leverages an off-the-shelf image text spotter with a strong customized tracker. \textbf{2)} We introduce the rescoring mechanism and long-short term matching module to adapt image text spotter to video datasets and enhance the tracker's capabilities. \textbf{3)} We establish the ArTVideo test set for addressing the absence of curved texts in current video datasets and evaluating the text spotters on videos with arbitrary-shape text. Extensive experiments on public challenging datasets and the established ArTVideo test set demonstrate the effectiveness of our baseline and its outstanding performance with less training budgets. 
For example, GoMatching achieves the highest ranking on ICDAR15-video and DSText. Especially on bilingual dataset BOVText, GoMatching obtains a 45\% improvement on MOTA compared to the recorded best performance~\cite{Wu2022CoText}.
On curved text dataset ArTVideo, GoMatching also surpasses previous SOTA method~\cite{Wu2022TransDETR} by a substantial margin.

\section{Related Works}
\subsection{Multi-Object Tracking}
Multi-object tracking methods follow the tracking-by-detection (TBD) or tracking-by-query-propagation (TBQP) pipeline.
TBD methods~\cite{wang2020towards,aharon2022bot,zhang2022bytetrack} employ detectors for localization and then use association algorithms to get object trajectories. Different from extending tracks frame-by-frame, GTR~\cite{zhou2022global} proposes to generate entire trajectories at once in Transformer.
TBQP paradigm extends query-based object detectors\cite{carion2020end,zhu2020deformable} to tracking. MOTR~\cite{zeng2022motr} detects object locations and serially updates its tracking queries for detecting the same items in the following frames, achieving an end-to-end solution. However, MOTR suffers from optimization conflict between detection and association~\cite{zhang2023motrv2,yu2023motrv3}, resulting in inferior detection performance. For the VTS task which additionally involves text recognition, a naive way of training all modules end-to-end may also lead to optimization conflict. In contrast, we explore inheriting prior knowledge of text spotting from ITS models while focusing on the tracking task.

\subsection{Image Text Spotting}
Early approaches~\cite{liao2020mask,wang2021pan++,liu2021abcnet} crafted RoI-based modules to bridge text detection and recognition. However, these methods ignored one vital issue, \ie, the synergy problem between the two tasks. 
To overcome this dilemma, recent Transformer-based methods~\cite{kittenplon2022towards,liu2023spts,ye2023deepsolo++,huang2023estextspotter} get rid of the fetters of RoI modules, and chase a better representation for the two tasks. For example, DETR-based TESTR~\cite{zhang2022text} uses two decoders for each task in parallel. In contrast, DeepSolo~\cite{ye2023deepsolo} proposes a unified and explicit query form for the two tasks, without harnessing dual decoders. However, the above methods cannot perform tracking in the video. 

\subsection{Video Text Spotting}
Compared to ITS, existing SOTA VTS methods still rely on RoI for recognition. 
CoText~\cite{Wu2022CoText} adopts a lightweight text spotter with Masked-RoI, then uses several encoders to fuse features derived from the spotter, and finally feeds them to a tracking head with cosine similarity matching.
TransDETR~\cite{Wu2022TransDETR} performs detection and tracking under the MOTR paradigm and then uses Rotated-RoI to extract features for the subsequent recognizer.
They pursue training all modules in an end-to-end manner. In comparison, we explore how to efficiently turn a RoI-free ITS model into a VTS one. We reveal the probability of freezing off-the-shelf ITS part and focusing on tracking, thereby saving training budgets while reaching SOTA performance.

\section{Methodology}
\subsection{Overview}
\begin{figure*}[ht]
  \centering
    \includegraphics[width=1.0\textwidth]{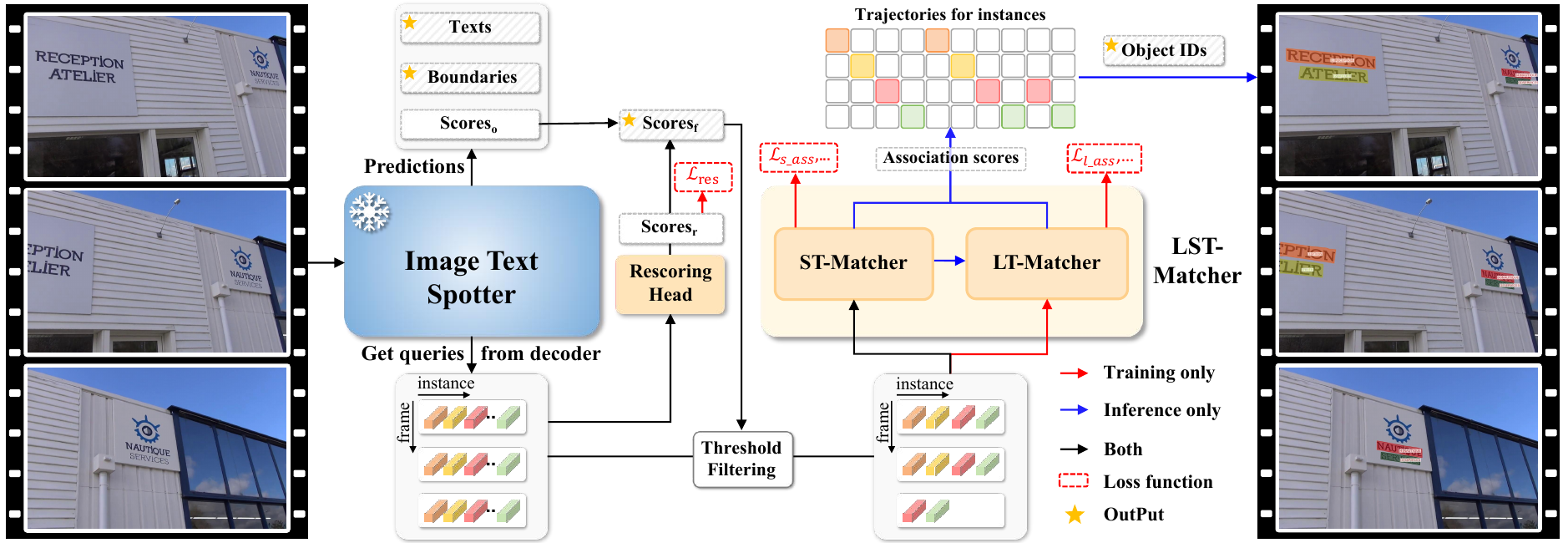}
    \caption{\textbf{The overall architecture of GoMatching}. 
    The frozen image text spotter provides text spotting results for frames. The rescoring mechanism considers both instance scores from the image text spotter and a trainable rescoring head to reduce performance degradation due to the domain gap. Long-short term matching module (LST-Matcher) assigns IDs to text instances based on the queries in long-short term frames. The yellow star sign `{\color{yellow}{$\bigstar$}}' indicates the final output of GoMatching.}
\label{fig:2}
\vspace{-5mm}
\end{figure*}

The architecture of GoMatching is presented in Fig.~\ref{fig:2}. It consists of a frozen image text spotter, a rescoring head, and a Long-Short Term Matching module (LST-Matcher).
We adopt an outstanding off-the-shelf image text spotter (\textit{i.e.}, DeepSolo) and freeze its parameters, with the aim of introducing strong text spotting capability into VTS while significantly reducing training cost.
In DeepSolo, there are $p$ sequences of queries used for final predictions, with each storing comprehensive semantics for a text instance.
To alleviate spotting performance degradation caused by the image-video domain gap, we devise a rescoring mechanism, which determines the confidence scores for text instances by considering both the scores from the image text spotter and a new trainable rescoring head. 
Finally, we design LST-Matcher to generate instance trajectories by leveraging long-short term information.

\subsection{Rescoring Mechanism}
Owing to the domain gap between image and video datasets, employing a frozen image text spotter for direct prediction may result in relative low recall due to low text confidence, further leading to a reduction in end-to-end spotting performance.
To ease this issue, we devise a rescoring mechanism via a lightweight rescoring head and a simple score fusion operation. Specifically, the rescoring head is designed to recompute the score for each query from the decoder in the image text spotter. It consists of a simple linear layer and is initialized with the parameters of the image text spotter's classification head.
The score fusion operation then decides the final scores by considering both the scores from the image text spotter and the rescoring head.
Let $C^{t}_o=\{c^{t}_{o_1},...,c^{t}_{o_p}\}$ be a set of original scores produced by image text spotter in frame $t$. $C^{t}_r=\{c^{t}_{r_1},...,c^{t}_{r_p}\}$ is a set of recomputed scores obtained from the rescoring head. We obtain the maximum value for each query as the final score, denoted as $C^{t}_f=\{c^{t}_{f_1}=max(c^{t}_{o_1}, c^{t}_{r_1}),...,c^{t}_{f_p}=max(c^{t}_{o_p}, c^{t}_{r_p})\}$. \
With final scores, the queries in frames are filtered by a threshold before being sent to LST-Matcher for association. 

\subsection{Long-Short Term Matching Module}

\begin{wrapfigure}{r}{0.5\textwidth}
\vspace{-10mm}
\begin{minipage}{0.5\textwidth}
\centering
\includegraphics[width=\textwidth]{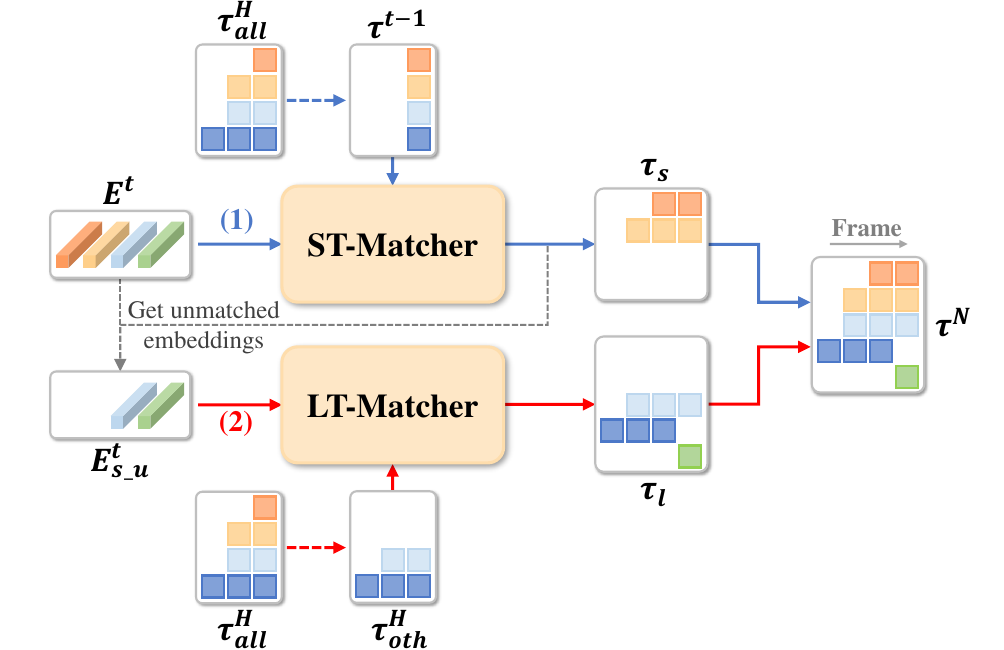} 
\end{minipage}
\caption{The inference pipeline of LST-Matcher, which is a two-stage association process: (1) ST-Matcher associates the instances with trajectories in previous frames as denoted by \textcolor[rgb]{0.3,0.5,1.0}{blue} lines. (2) LT-Matcher associates the remaining unmatched instances by utilizing other trajectories in history frames as denoted by {\color{red}{red}} lines.}
 \label{fig:3}
\end{wrapfigure}

Long-short term matching module (LST-Matcher) consists of two sub-modules: the Short Term Matching module (ST-Matcher) and the Long Term Matching module (LT-Matcher), which own the same structure.
ST-Matcher is steered to match simple instances between adjacent frames into trajectories, while LT-Matcher is responsible for using long term information to address the unmatched instances due to severe occlusions or strong appearance changes.
Each of them contains a one-layer Transformer encoder and a one-layer Transformer decoder~\cite{zhou2022global}. We use a simple multi-layer perceptron (MLP) to map the filtered text instance queries into embeddings as the input, getting rid of using RoI features as in most existing MOT methods. In the encoder, historical embeddings are enhanced by self-attention. The decoder takes embeddings in the current frame as query and enhanced historical embeddings as key for cross-attention, and computes the association score matrix. The current instances are then linked to the existing trajectories composed of historical embeddings or generate new trajectories according to the association score matrix.

To be specific, supposing a given clip including $T$ frames and $N_t$ text instances in frame $t$ after threshold filtering. 
$Q^t = \{q^{t}_{1},...,q^{t}_{N_t}\}$ is the set of text instance queries in frame $t$.
Initially, we use a two-layer MLP to map these frozen queries into embeddings $E^t = \{e^{t}_{1},...,e^{t}_{N_t}\}$. The set of embeddings in all frames is denoted as $E^L = E^1 \cup...\cup E^T$. Let the universal set of embeddings in adjacent frames of the input batch be denoted as $E^S = E^{S_2} \cup E^{S_3} \cup... \cup E^{S_T}$ and $E^{S_t} = E^{t-1} \cup E^t$.
Based on the predictions of image text spotter, we obtain their corresponding bounding boxes $B^{t} = \{b^{t}_{1},...,b^{t}_{N_t}\}$.
Let $\tau = \{\tau_1,...,\tau_K\}$ be the set of ground-truth (GT) trajectories of all instances in the clip, where $\tau_k = \{\tau^1_k,...,\tau^{T}_k\}$ describes a tube of instance locations $\tau^t_k \in \mathbbm{R}^4\cup\{\emptyset\}$ through time. $\tau^t_k = \emptyset$ means the absence of instance $k$ in frame $t$. Let $\hat{\alpha}^t_k$ be the matched instance index for $\tau^t_k$ according to the following equation:
\begin{equation}
    \hat{\alpha}^t_k = \left\{
    \begin{array}{ll}
         \emptyset, \quad \text{if } \tau^t_k=\emptyset \text{ or max}_i (\textit{IoU}(b^{t}_{i},\tau^t_k)) < 0.5 \\
         \text{argmax}_i (\textit{IoU}(b^{t}_{i},\tau^t_k)), \quad\text{otherwise}
    \end{array}.
    \right.
    \label{eq:1}
\end{equation}
ST-Matcher calculates a short-term 
association score $v^t_i(e^t_{\hat{\alpha}^t_k}, E^{S_t})  \in \mathbb{R}^{N_{S_t}}$ for $i$-th instances in frame $t$, where $e^t_{\hat{\alpha}^t_k} \in \mathbb{R}^D$ is a trajectory query and $N_{S_t} = N_t + N_{t-1}$.
LT-Matcher calculates a long-term trajectory-specific association score $u^t_i(e_k, E^L) \in \mathbb{R}^{N}$ for $i$-th instances in frame $t$, where $e_k \in \{e^1_{\hat{\alpha}^1_k},e^2_{\hat{\alpha}^2_k},...,e^T_{\hat{\alpha}^T_k}\}, N = \sum^T_{t=1}N_t$.
Specifically, when $v^t_i(e^t_{\hat{\alpha}^t_k}, E^{S_t}) = 0$ and $u^t_i(e_k, E^L) = 0$, it means no association at time $t$. Then, ST-Matcher and LT-Matcher can predict distributions of short-term and long-term associations for all instance $i$ in frame $t$ which can be written as:
\begin{equation}
    P_{s_a}(e^t_{\hat{\alpha}^t_k},E^{S_t})=\frac{\exp(v^t_i(e^t_{\hat{\alpha}^t_k}, E^{S_t}))}{\sum_{j\in\{\emptyset,1,...,N_t\}}\exp(v^t_j(e^t_{\hat{\alpha}^t_k}, E^{S_t}))} ,
    \label{eq:2}
\end{equation}
\begin{equation}
    P_{l_a}(e_k,E^L)=\frac{\exp(u^t_i(e_k, E^L))}{\sum_{j\in\{\emptyset,1,...,N_t\}}\exp(u^t_j(e_k, E^L))}  .
    \label{eq:3}
\end{equation}

To ensure sufficient training of ST-Matcher and LT-Matcher, embeddings set $E^S$ and $E^L$ are fed into ST-Matcher and LT-Matcher during training, respectively.

During inference, we engage a memory bank to store the instance trajectories from $H$ history frames for long term association.
All filtered instances in each frame are further processed by non-maximum-suppression (NMS) before being fed into LST-Matcher for association.
Unlike the training phase, where ST-Matcher and LT-Matcheder are independent of each other, LST-Matcher comprises a two-stage associating procedure as described in Fig.~\ref{fig:3}. Concretely, ST-Matcher first matches the embedding $E^t$ in the current frame $t$ with the trajectories $\tau^{t-1}$ in the previous frame $t-1$.
Then, LT-Matcher employs other trajectories $\tau^H_{oth}$ in the memory bank to associate the unmatched ones $E^t_{s\_u}$ with low association score in ST-Matcher caused by the heavy occlusion or strong appearance changes. 
If the association score with any trajectory calculated in ST-Matcher or LT-Matcher is higher than a threshold $\theta$, the instance is linked to the trajectory with the highest score. 
Otherwise, this instance is used to initiate a new trajectory.
Finally, we combine the trajectories $\tau_s$ and $\tau_l$ predicted by ST-Matcher and LT-Matcher to obtain new trajectories $\tau^N$ for tracking in the next frame.

\subsection{Optimization}
\textbf{Rescoring Loss.} To train the rescoring head, we following DETR~\cite{carion2020end} and use Hungarian algorithm~\cite{kuhn1955hungarian} to find a bipartite matching $\hat{\sigma}$ between the prediction set $\hat{Y}$ and the ground truth set $Y$ with minimum matching cost $\mathcal{C}$:
\begin{equation}
    \hat{\sigma} =  \arg\min\limits_{\sigma}\sum^{N}_{i}\mathcal{C}(Y_i, \hat{Y}_{\sigma(i)}),
    \label{eq:4}
\end{equation}
where \textit{N} is the number of ground truth instances per frame. The cost $\mathcal{C}$ can be defined as:
\begin{equation}
    \mathcal{C}(Y_i, \hat{Y}_{\sigma(i)}) = \lambda_c\mathcal{L}_{cls}(\hat{p}_{\sigma(i)}(c_i)) + \lambda_b\sum^N_1\Vert b_i - \hat{b}_{\sigma(i)}\Vert,
    \label{eq:5}
\end{equation}
where $\lambda_c$ and $\lambda_b$ serve as the hyper-parameters to balance different tasks. $\hat{p}_{\sigma(i)}(c_i)$ and $\hat{b}_{\sigma(i)}$ are the probability for ground truth class $c_i$ and the predicition of bounding box respectively, and $b_i$ represents the ground truth bounding box. $\mathcal{L}_{cls}$ is the focal loss~\cite{lin2017focal}. Specifically, the focal loss for training the rescoring head can be formulated as:
\vspace{-6pt}
\begin{equation}
    \mathcal{L}_{res} = \sum^N_1 [-\mathds{1}_{\{c_i\neq\varnothing\}}\alpha(1 - \hat{p}_{\hat{\sigma}(i)}(c_i))^\gamma\log(\hat{p}_{\hat{\sigma}(i)}(c_i)) -\mathds{1}_{\{c_i=\varnothing\}}(1-\alpha)(\hat{p}_{\hat{\sigma}(i)}(c_i))^\gamma\log(1-\hat{p}_{\hat{\sigma}(i)}(c_i))],
    \label{eq:6}
\end{equation}
where $\alpha$ and $\gamma$ are the hyper-parameters of focal loss.

\noindent\textbf{Long-Short Association Loss.} In ST-Matcher, we only consider each trajectory in the universal set of adjacent frames, while in LT-Matcher we consider each trajectory in all long term frames. For each trajectory, we optimize the log-likelihood of its assignments $\hat{\alpha}_k$ following GTR~\cite{zhou2022global}:
\vspace{-4pt}
\begin{equation}
    \mathcal{L}_{s\_ass}(E^S,\hat{\tau}_k)=-\sum^{T}_{t=2}\log P_{s_a}(\hat{\alpha}^t_k|e^t_{\hat{\alpha}^t_k}, E^{S_t}),
    \label{eq:7}
\end{equation}
\begin{equation}
    \mathcal{L}_{l\_ass}(E^L,\hat{\tau}_k)=-\sum_{w}\sum^{T}_{t=1}\log P_{l_a}(\hat{\alpha}^t_k|E^w_{\hat{\alpha}^w_k}, E^{L}),
    \label{eq:8}
\end{equation}
where $w\in\{1,...,T | \hat{\alpha}^w_k\neq\emptyset\}$.

In ST-Matcher and LT-Matcher, empty trajectories would be generated for these unassociated queries, and their optimization goals can be defined as:
\vspace{-4pt}
\begin{equation}
    \mathcal{L}_{s\_bg}(E^S)=-\sum_{j:\nexists\hat{\alpha}^t_k=j}\sum^{T}_{t=2}\log P_{s_a}(\alpha^t=\emptyset|e^t_j, E^{S_t}),
    \label{eq:9}
\end{equation}
\vspace{-4pt}
\begin{equation}
    \mathcal{L}_{l\_bg}(E^L)=-\sum^{T}_{w=1}\sum_{j:\nexists\hat{\alpha}^w_k=j}\sum^{T}_{t=1}\log P_{l_a}(\alpha^t=\emptyset|E^w_j, E^{L}).
    \label{eq:10}
\end{equation}

\vspace{-6pt}
Finally, we can get the long-short association loss as follows:
\begin{equation}
    \mathcal{L}_{asso} = \mathcal{L}_{s\_bg} + \mathcal{L}_{l\_bg} + \sum_{\hat{\tau}_k}(\mathcal{L}_{s\_ass} + \mathcal{L}_{l\_ass}).
    \label{eq:11}
\end{equation}

\vspace{-6pt}
\noindent\textbf{Overall Loss.} Combined with the rescore loss $\mathcal{L}_{res}$ in Eq. (\ref{eq:6}) and the long-short association loss $\mathcal{L}_{asso}$ in Eq. (\ref{eq:11}), the final training loss can be defined as:
\begin{equation}
    \mathcal{L} = \lambda_{res}\mathcal{L}_{res} + \lambda_{asso}\mathcal{L}_{asso},
    \label{eq:12}
\end{equation}
where the hyper-parameters $\lambda_{res}$ and $\lambda_{asso}$ are the weights of $\mathcal{L}_{res}$ and $\mathcal{L}_{asso}$, respectively.

\section{Experiments}
\subsection{Datasets and Evaluation Metrics}
\label{exp: 'dataset'}
\textbf{ICDAR15-video}~\cite{Karatzas2015ICDAR15} is a word-level video text reading benchmark annotated with quadrilateral bounding boxes, comprising a training set of 25 videos and a test set of 24 videos. It focuses on wild scenarios, such as driving on the road, exploring shopping streets, walking in a supermarket, \textit{etc.}

\noindent\textbf{BOVText}~\cite{Wu2021BOVText} is a large-scale, bilingual, and open-world benchmark for video text spotting, encompassing English and Chinese. The dataset is meticulously collected from \textit{YouTube} and \textit{KuaiShou} with different scenarios.
The text box annotations are represented as quadrilaterals at the textline level.

\noindent\textbf{DSText}~\cite{Wu2023DSText} is a newly proposed dataset, and focuses on dense and small text reading challenges in the video. This dataset provides 50 training videos and 50 test videos. Compared with the previous datasets, DSText mainly includes the following three new challenges: dense video texts, high-proportioned small texts, and various new scenarios, \textit{e.g.}, `Game', `Sports', \textit{etc.} Similar to ICDAR15-video, DSText adopts word-level annotations, which are labeled with quadrilaterals.

\noindent\textbf{ArTVideo} is a novel word-level test set established in this work to evaluate the performance of arbitrary-shaped video text, which is absent in the VTS community. It contains 20 videos with about 30\% curved text instances. Straight text is annotated with quadrilaterals, while curved text is annotated with polygons. More details are provided in the Appendix \ref{appx: artvideo}.

\noindent\textbf{Evaluation Metrics.} To evaluate performance, we adopt three evaluation metrics commonly used in ICDAR15-video competition and DSText competition, including MOTA~\cite{bernardin2008evaluating}, MOTP, and IDF1~\cite{ristani2016performance}.

\subsection{Implementation Details}
In all experiments, we only use a single NVIDIA GeForce RTX 3090 (24G) GPU to train and evaluate GoMatching. As for the image text spotter in GoMatching, we apply the officially released DeepSolo~\cite{ye2023deepsolo}. During fine-tuning GoMatching on downstream video datasets, we only update the rescoring head and LST-Matcher, while keeping DeepSolo frozen. 
More inference settings can be seen in the Appendix \ref{appx: settings}.

\noindent\textbf{Training Setting.} 
The text spotting part of GoMatching is initialized with off-the-shelf DeepSolo weights and kept frozen in all experiments. We optimize other modules on video datasets.
We follow EfficientDet~\cite{tan2020efficientdet} to adopt the scale-and-crop augmentation strategy with a resolution of 1280. The batch size $T$ is 6. All frames in a batch are from the same video. Text instances with fusion scores higher than 0.3 are fed into the LST-Matcher during training.
AdamW~\cite{loshchilov2017decoupled} is used as the optimizer. We adopt the warmup cosine annealing learning rate strategy with the initial learning rate being set to 0.00005. 
The loss weights $\lambda_{res}$ and $\lambda_{asso}$ are set to 1.0 and 0.5, respectively. For focal loss, $\alpha$ is 0.25 and $\gamma$ is 2.0 as in~\cite{carion2020end,ye2023deepsolo}. The model is trained for 30k iterations on all downstream video datasets.

\subsection{Comparison with State-of-the-art Methods}
\textbf{Results on ICDAR15-video.} To evaluate the effectiveness of GoMatching on oriented video text, we conduct a comparison with the state-of-the-art methods on ICDAR15-video presented in Table~\ref{table:1}. As can be seen, GoMatching ranks first in all metrics on the ICDAR15-video leaderboard. 
By effectively combining a robust image text spotter with a strong tracker, GoMatching improves the best performance by 5.08\% MOTA, 0.75\% MOTP, and 3.16\% IDF1, respectively.
Furthermore, owing to the substantial enhancement in recognition and tracking capabilities (details can be found in Sec. \ref{ablation} and Appendix \ref{appx: more vis}), GoMatching outperforms the current SOTA single-model method TransDETR by 11.08\% MOTA, 3.92\% MOTP, and 7.31\% IDF1, respectively.

\noindent\textbf{Results on BOVText.} Except for the English word recognition scenario, GoMatching can readily adapt to other video text recognition scenarios, such as Chinese text line recognition. For BOVText, which focuses on English and Chinese textline recognition, we employ the DeepSolo trained on bilingual textline datasets and then fine-tune GoMatching on BOVText. The results are presented in Table~\ref{table:3}. It is evident that GoMatching achieves a new record on the BOVText dataset and surpasses previous methods significantly. GoMatching exhibits superior performance over the previous SOTA method CoText~\cite{Wu2022CoText}, with improvements of 41.5\% on MOTA, 6.9\% on MOTP, and 14.3\% on IDF1. Such exceptional performance of GoMatching on BOVText suggests its proficiency in spotting both Chinese and English text in videos. Moreover, it can be easily extended to other languages by adapting the image spotter.

\noindent\textbf{Results on DSText.} We further conduct experiments on DSText with dense and small video text scenarios. Results are presented in Table~\ref{table:2}. It is worth noting that most of the previous methods on the DSText leaderboard used an ensemble of multiple models and
large public datasets to enhance their performance~\cite{Wu2023DSText}. For example, \textit{TencentOCR} integrates the detection results of DBNet~\cite{liao2020real} and Cascade MaskRCNN~\cite{cai2019cascade} built with multiple backbone architectures, combines the Parseq~\cite{bautista2022scene} text recognizer, and further improves the end-to-end tracking with ByteTrack~\cite{zhang2022bytetrack}. \textit{DA} adopts Mask R-CNN~\cite{he2017mask} and DBNet to detect text, then uses BotSORT~\cite{aharon2022bot} to replace the tracker in VideoTextSCM~\cite{gao2021video} and employs the Parseq model for recognition.
As a single model with a frozen image text spotter, GoMatching also shows competitive performance compared to other ensembling methods on the leaderboard. GoMatching ranks first (22.83\%) on MOTA, second (80.43\%) on MOTP, and third (46.09\%) on IDF1. Moreover, compared to the SOTA single-model method, GoMatching achieves substantial improvements of 45.46\% and 19.66\% on MOTA and IDF1, respectively.

\noindent\textbf{Results on ArTVideo.} We test TransDETR and GoMatching on ArTVideo to compare the zero-shot text spotting capabilities for arbitrary-shaped text. For a fair comparison, both TransDETR and GoMatching are trained on ICDAR15-video. Unlike ICDAR15-video and DSText which only have straight text, ArTVideo has a substantial number of curved text, so we report results under four settings: tracking results on both straight and curved text, spotting results on both straight and curved text, tracking results on curved text only, and spotting results on curved text only. As shown in Table~\ref{table:4}, GoMatching outperforms TransDETR under all settings. Especially when involving an additional recognition task (end-to-end spotting) or only considering curved text, the performance advantages of GoMatching are more significant. This further confirms that the previous SOTA methods have unsatisfactory recognition capabilities and limited adaptability to complex scenarios. Furthermore, as shown in Fig.~\ref{fig:1}(b), GoMatching achieves excellent performance while significantly reducing the training budget. 

\begin{figure*}[t]
\centering
\resizebox{\textwidth}{!}{
\begin{tabular}{cccc}
  \includegraphics[width=0.25\textwidth]{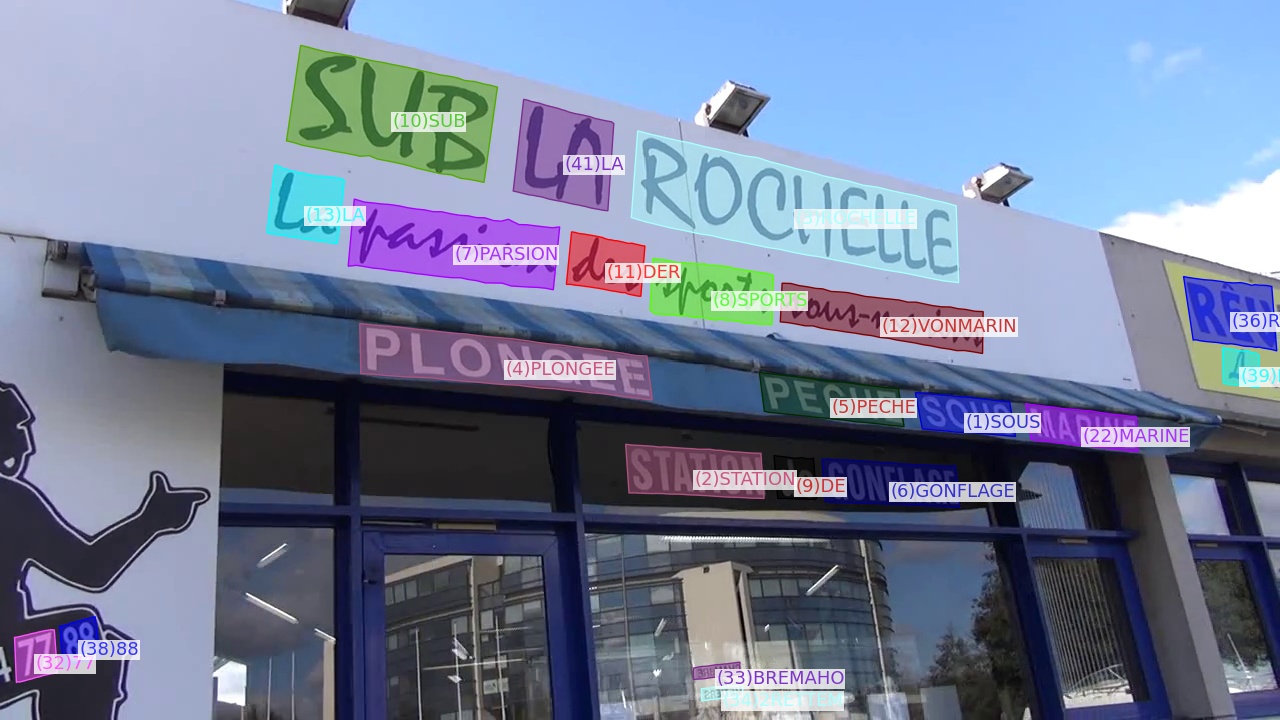} 
  &\includegraphics[width=0.25\textwidth]{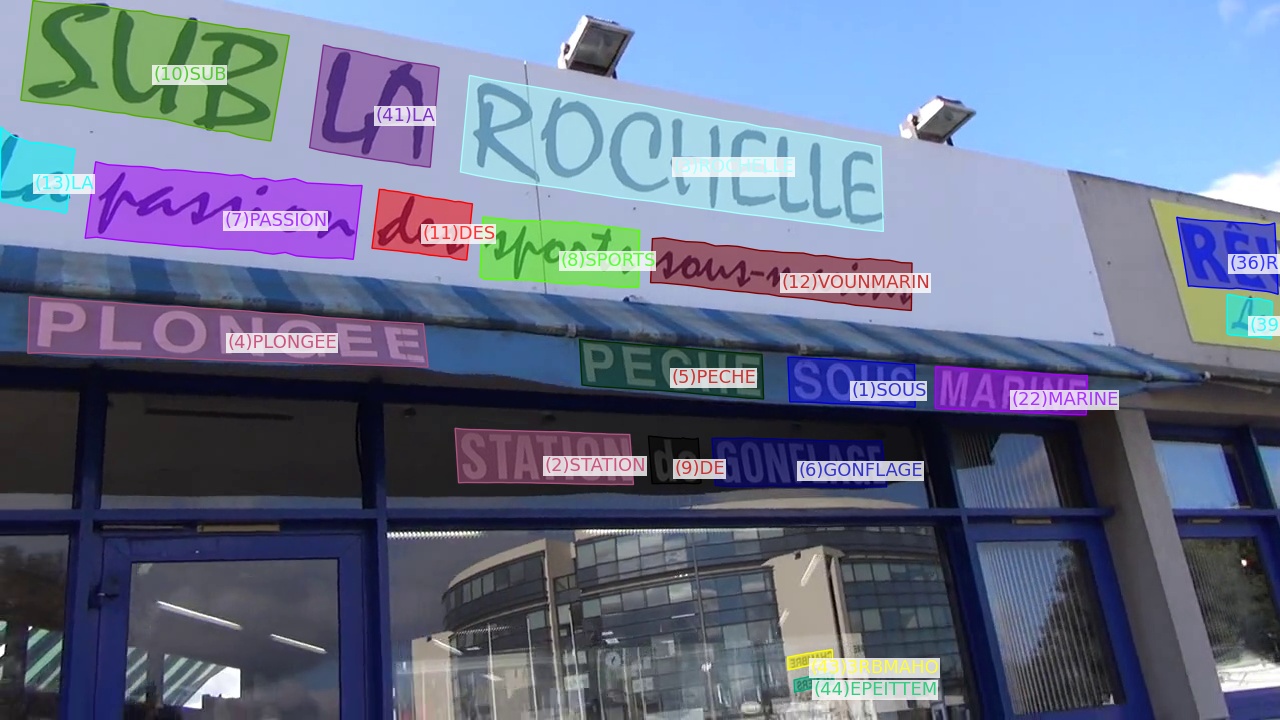}
  &\includegraphics[width=0.25\textwidth]{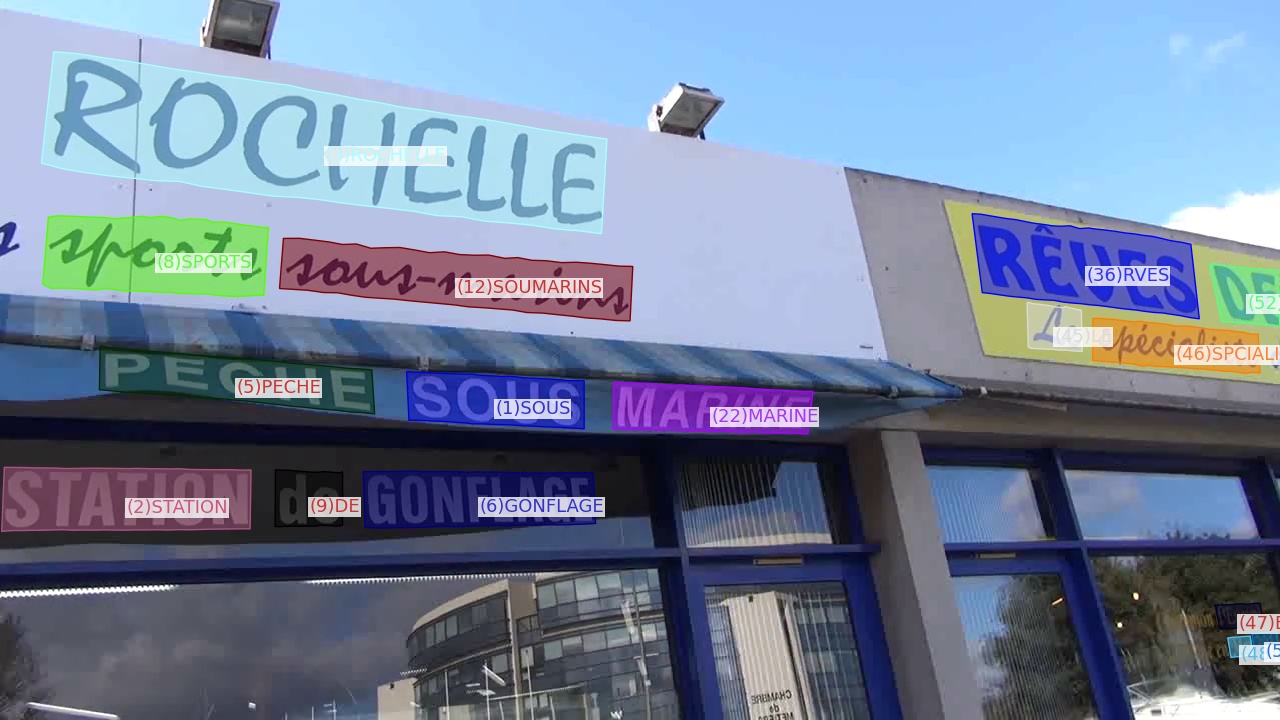}
  &\includegraphics[width=0.25\textwidth]{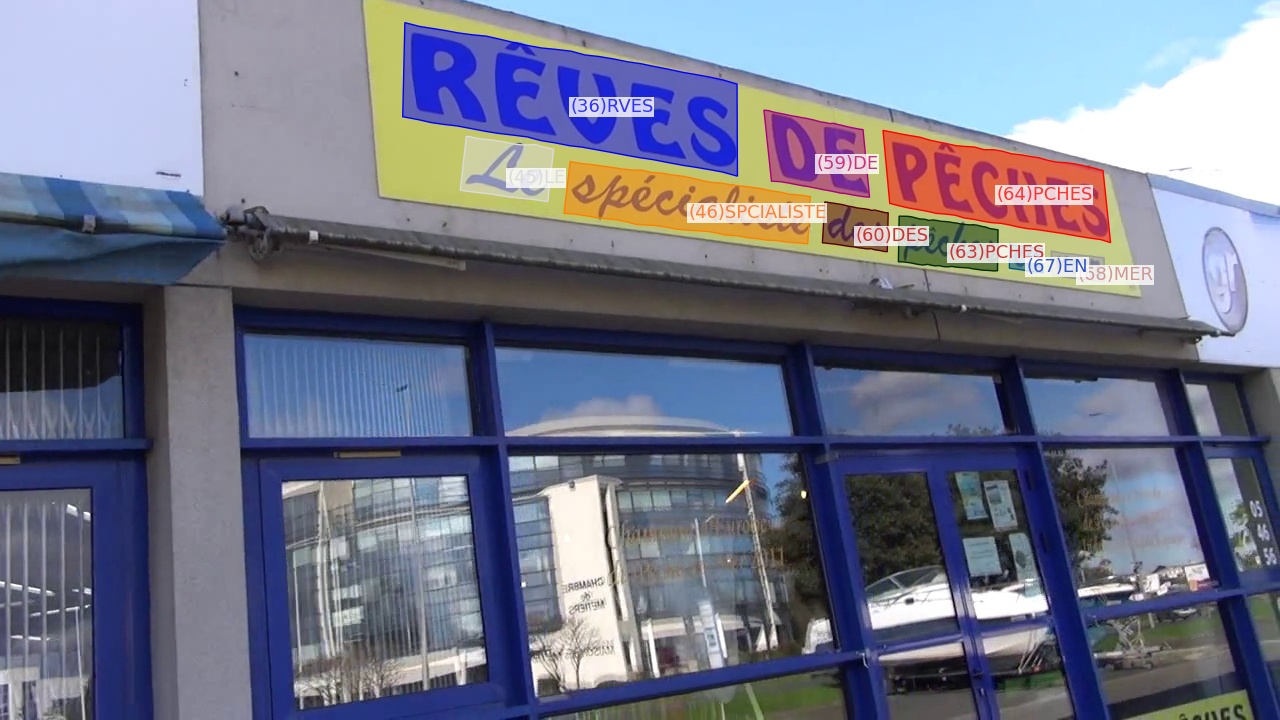}
  \\ 
  \includegraphics[width=0.25\textwidth]{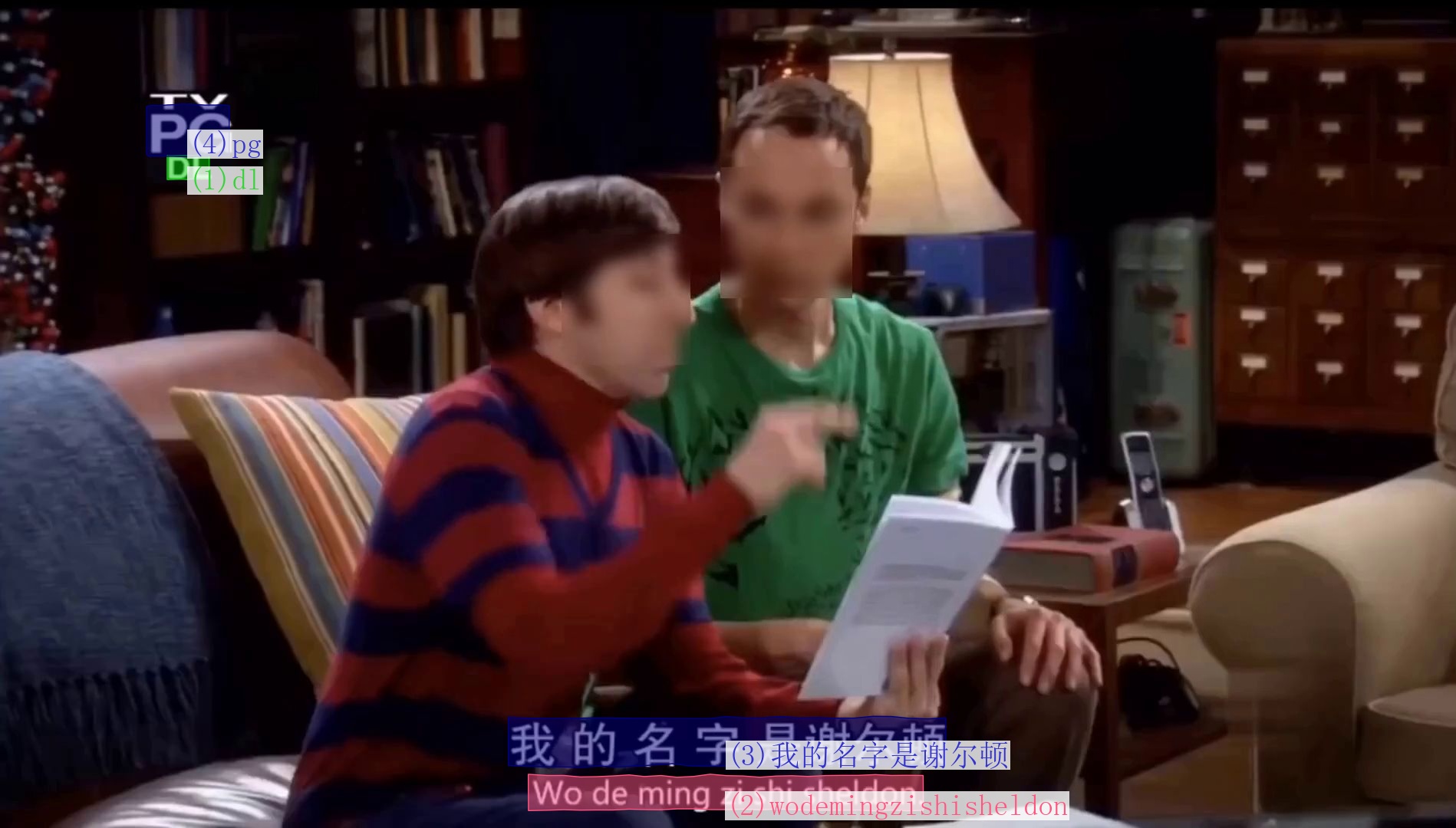} 
  &\includegraphics[width=0.25\textwidth]{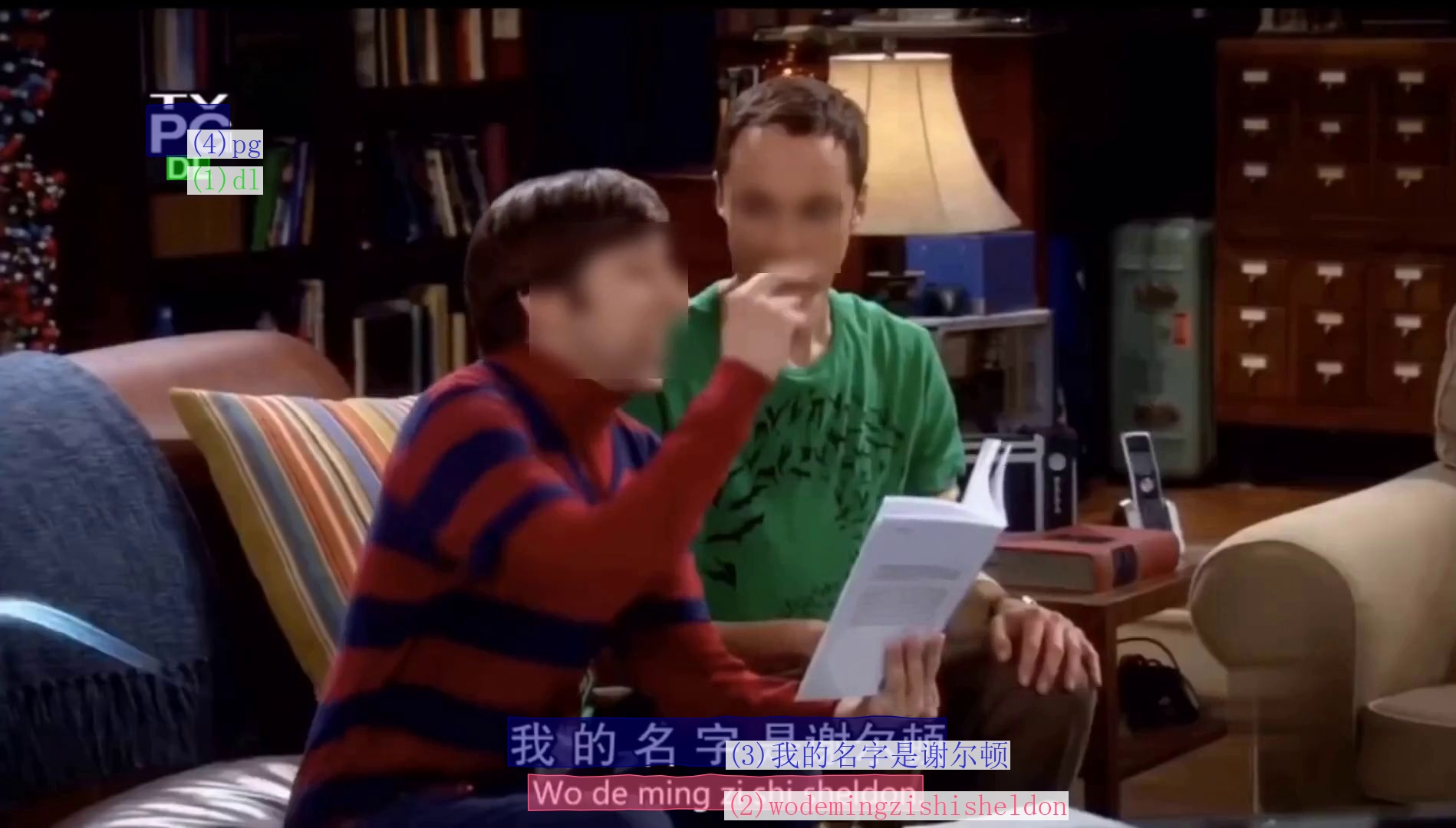}
  &\includegraphics[width=0.25\textwidth]{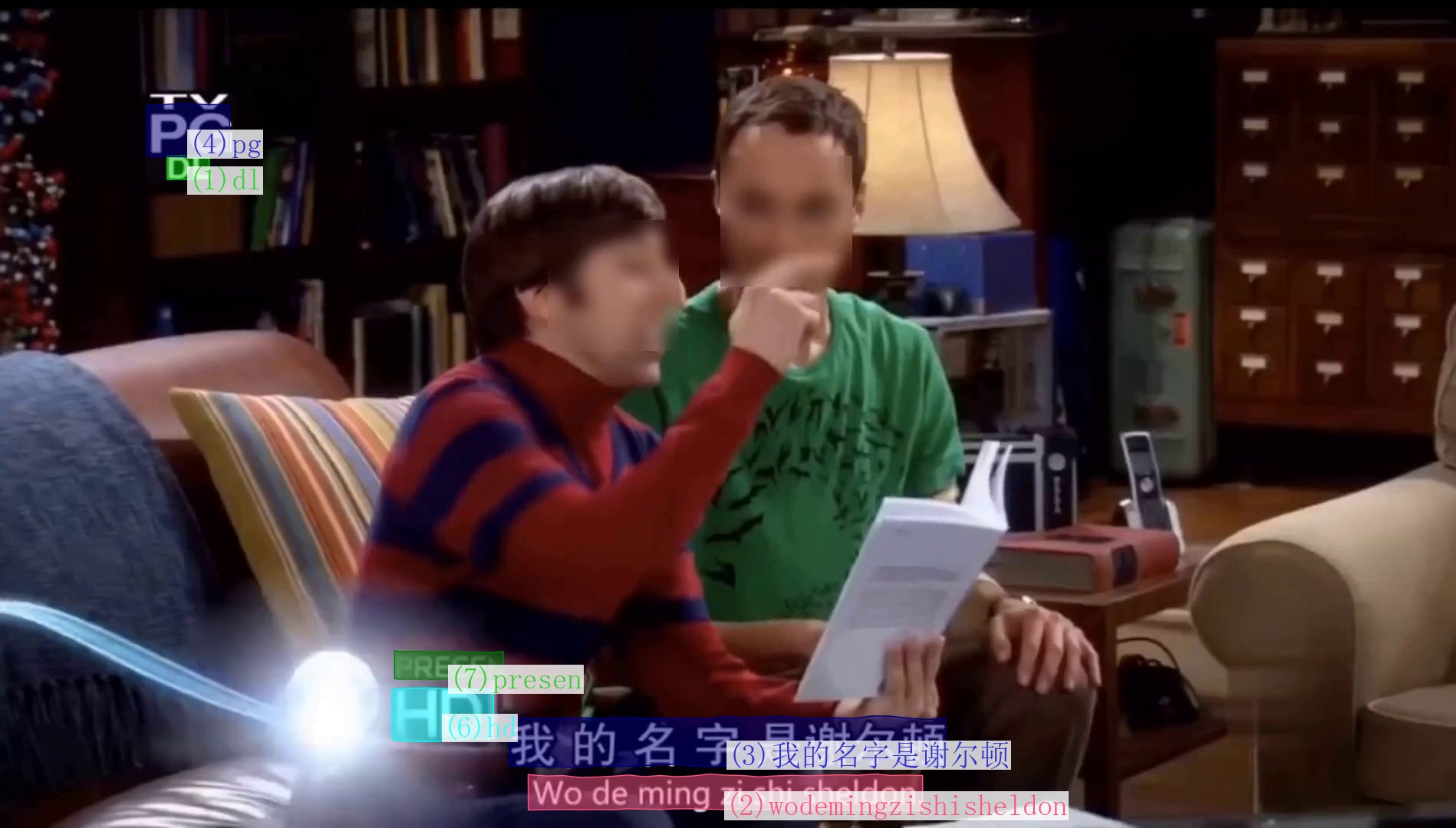}
  &\includegraphics[width=0.25\textwidth]{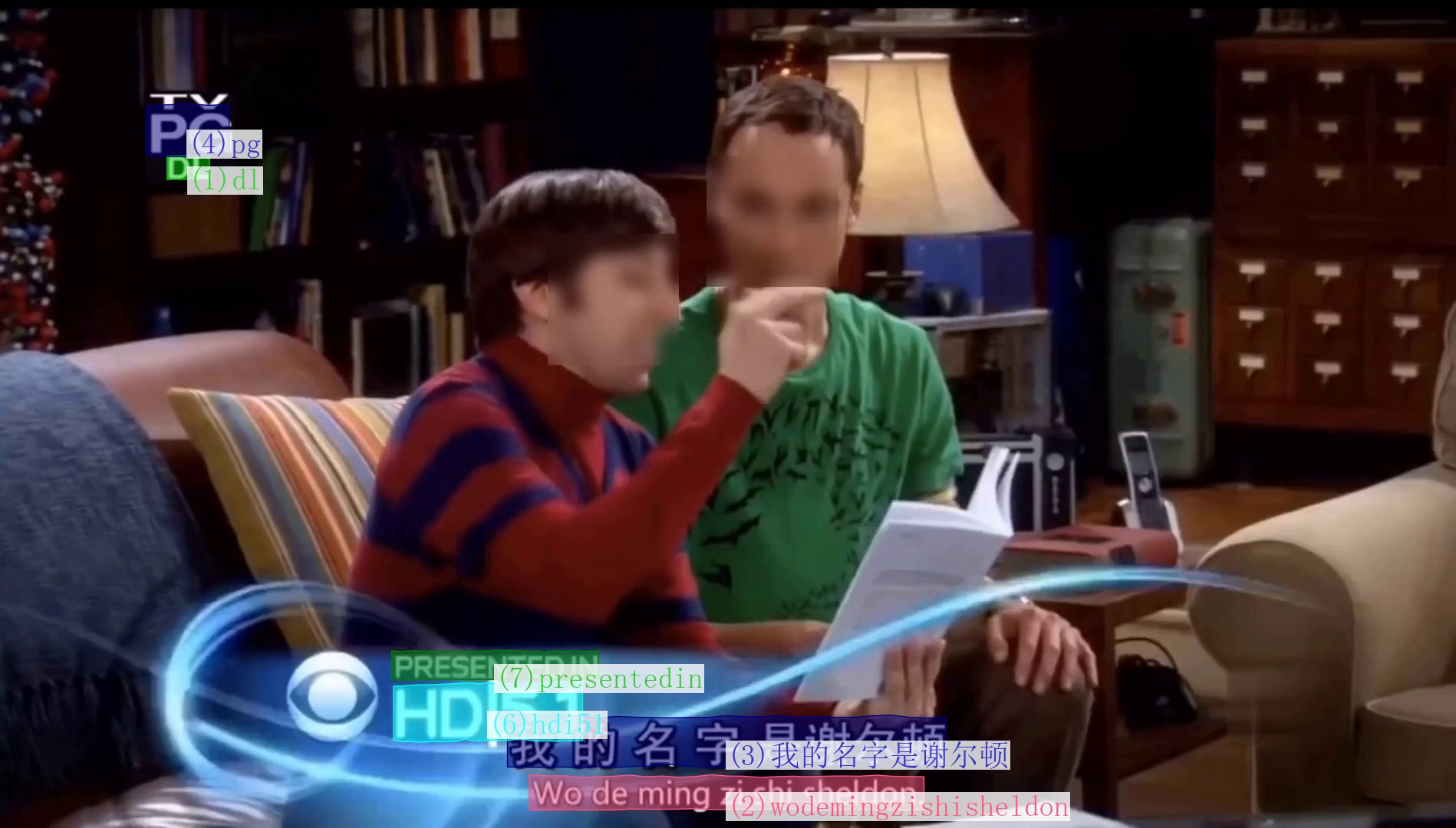}
  \\
  \includegraphics[width=0.25\textwidth]{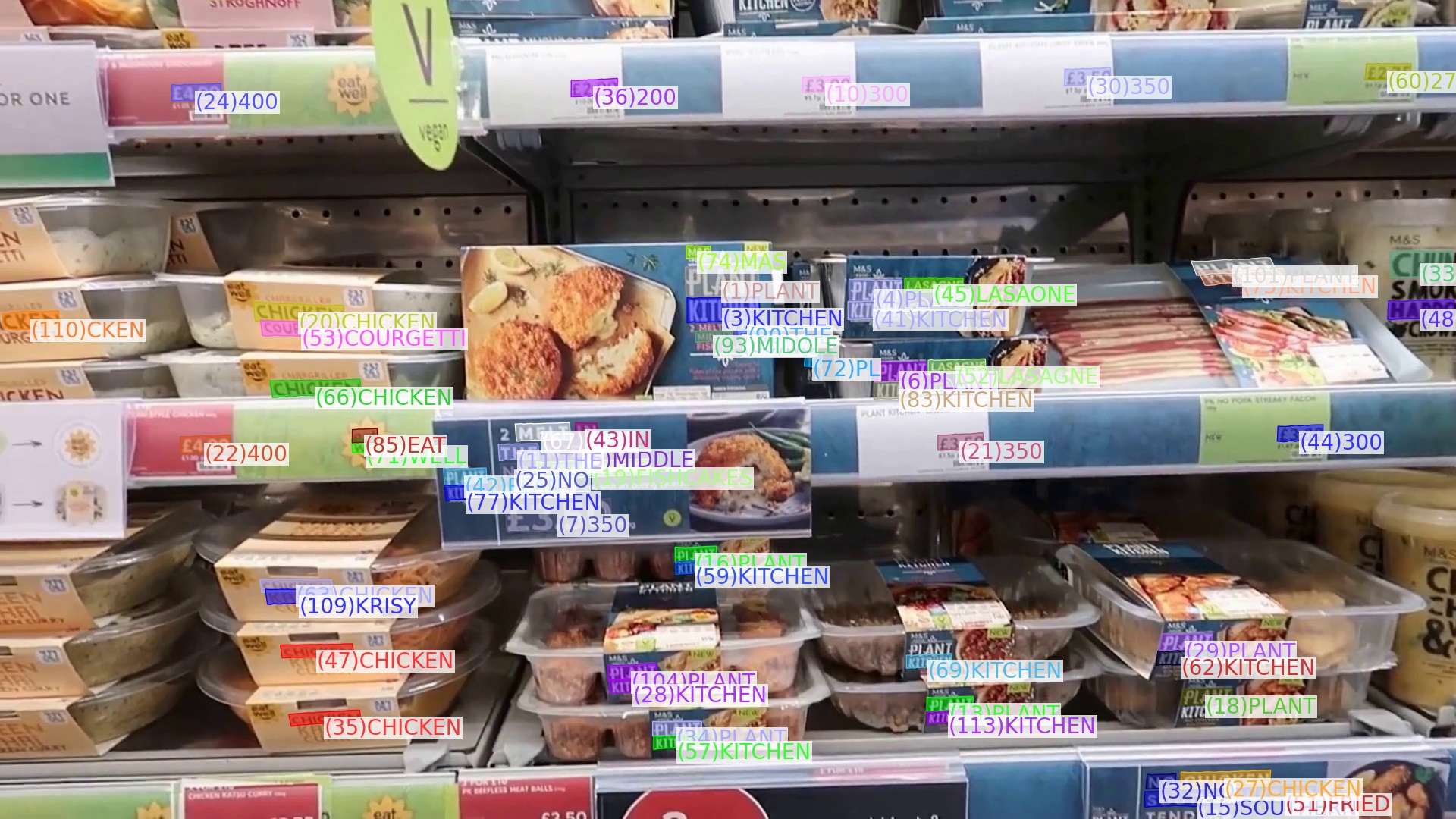}
  &\includegraphics[width=0.25\textwidth]{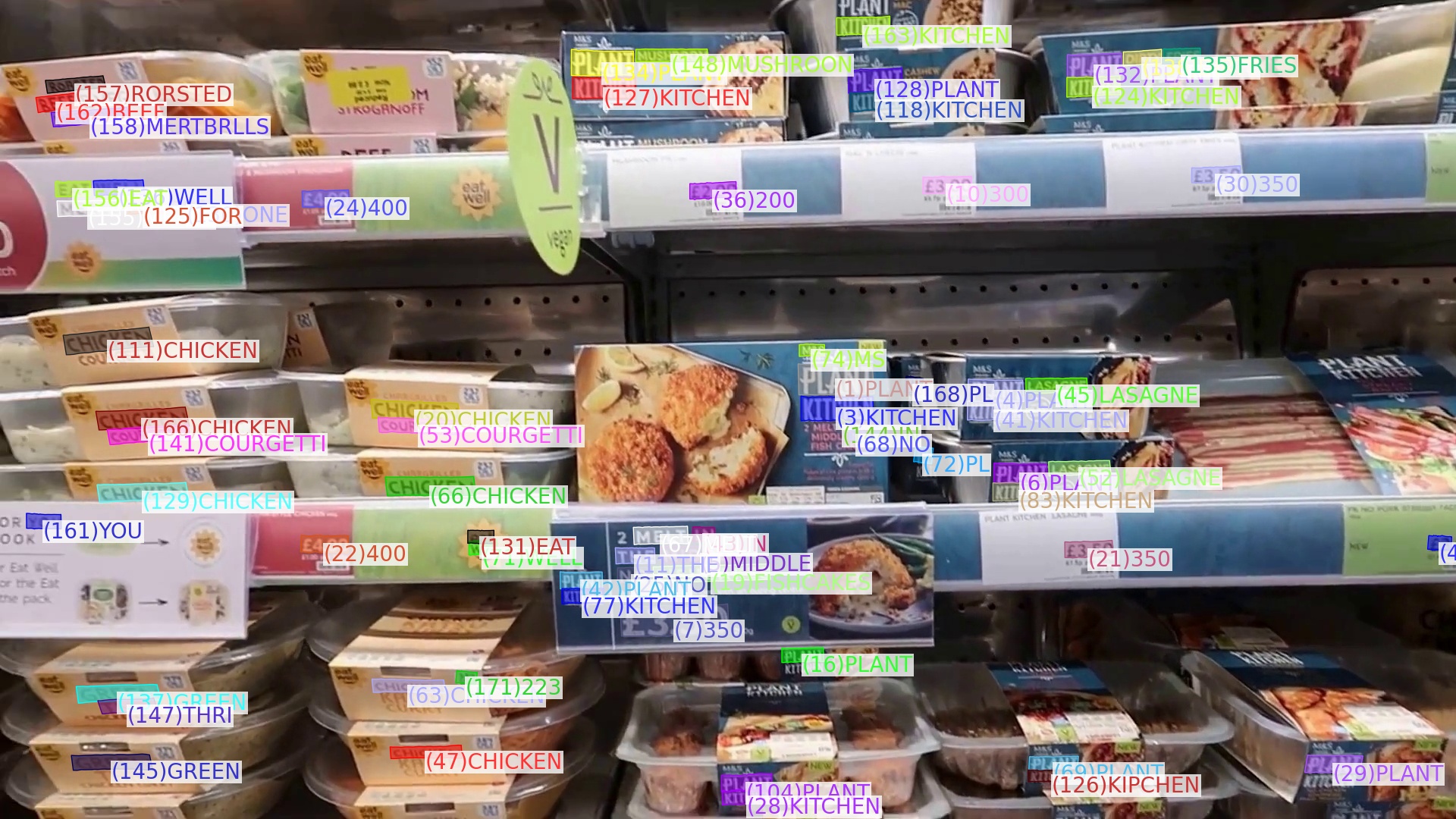}
  &\includegraphics[width=0.25\textwidth]{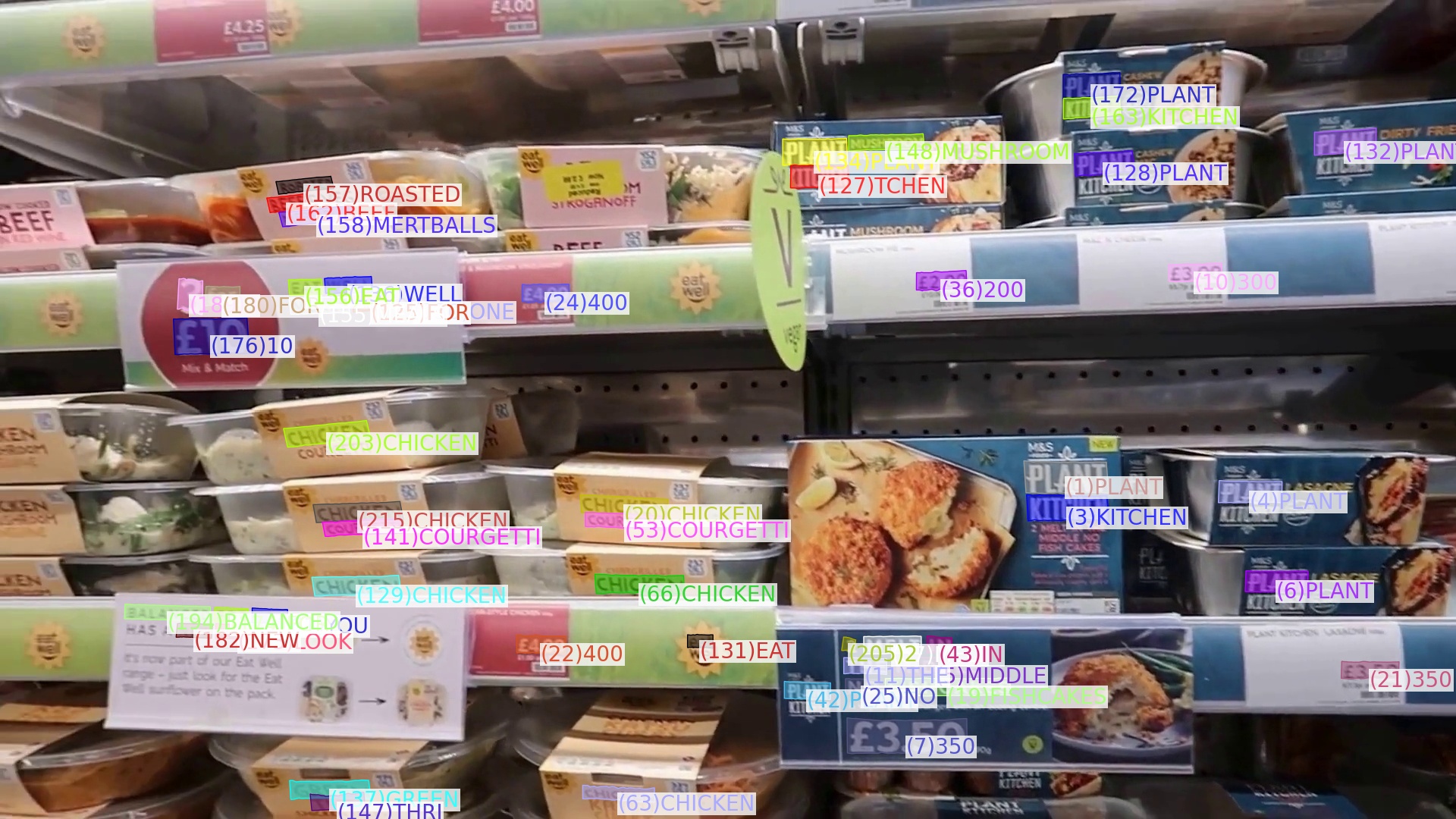}
  &\includegraphics[width=0.25\textwidth]{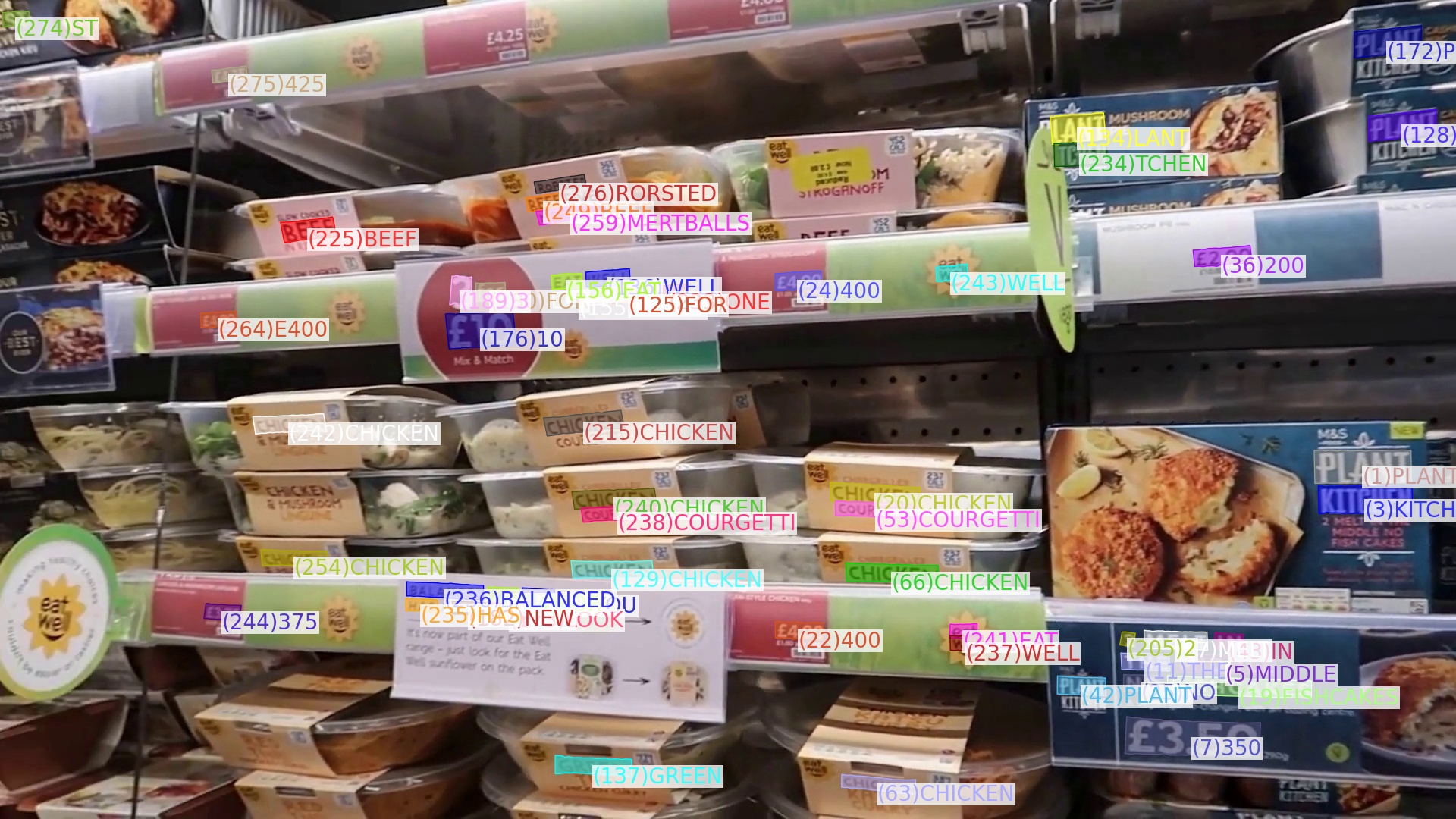}
  \\
  \includegraphics[width=0.25\textwidth]{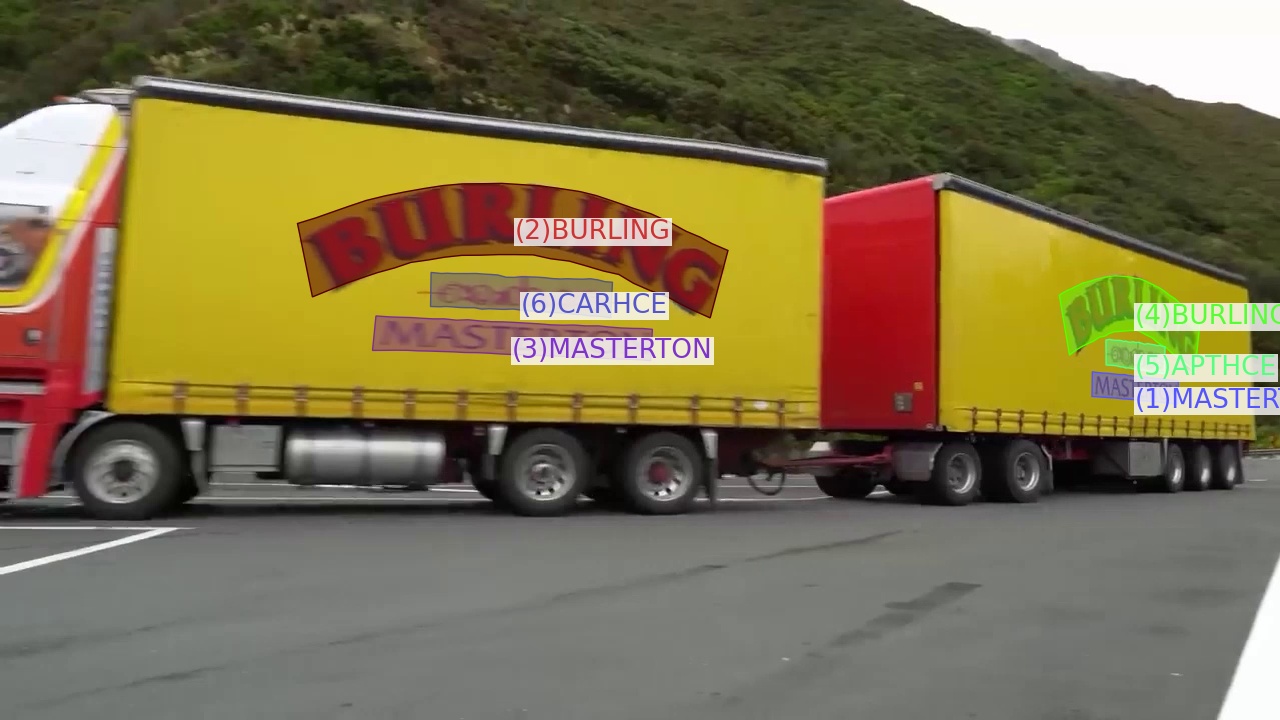}
  &\includegraphics[width=0.25\textwidth]{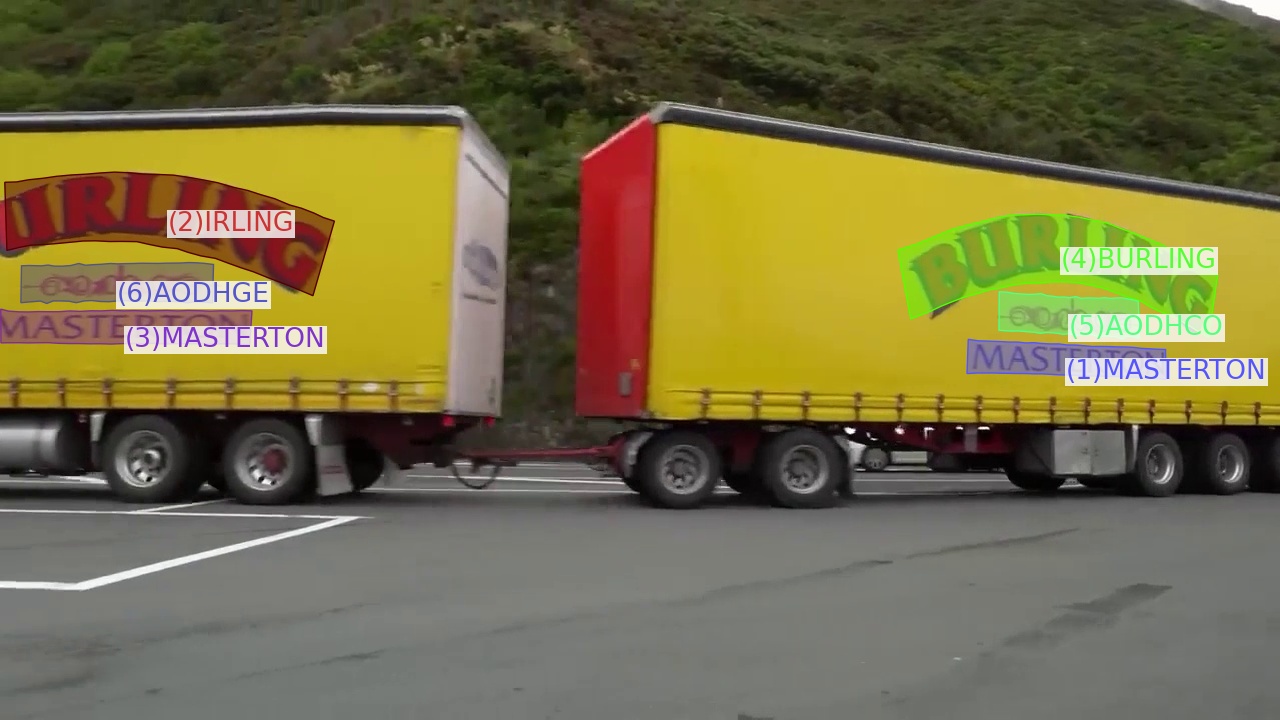}
  &\includegraphics[width=0.25\textwidth]{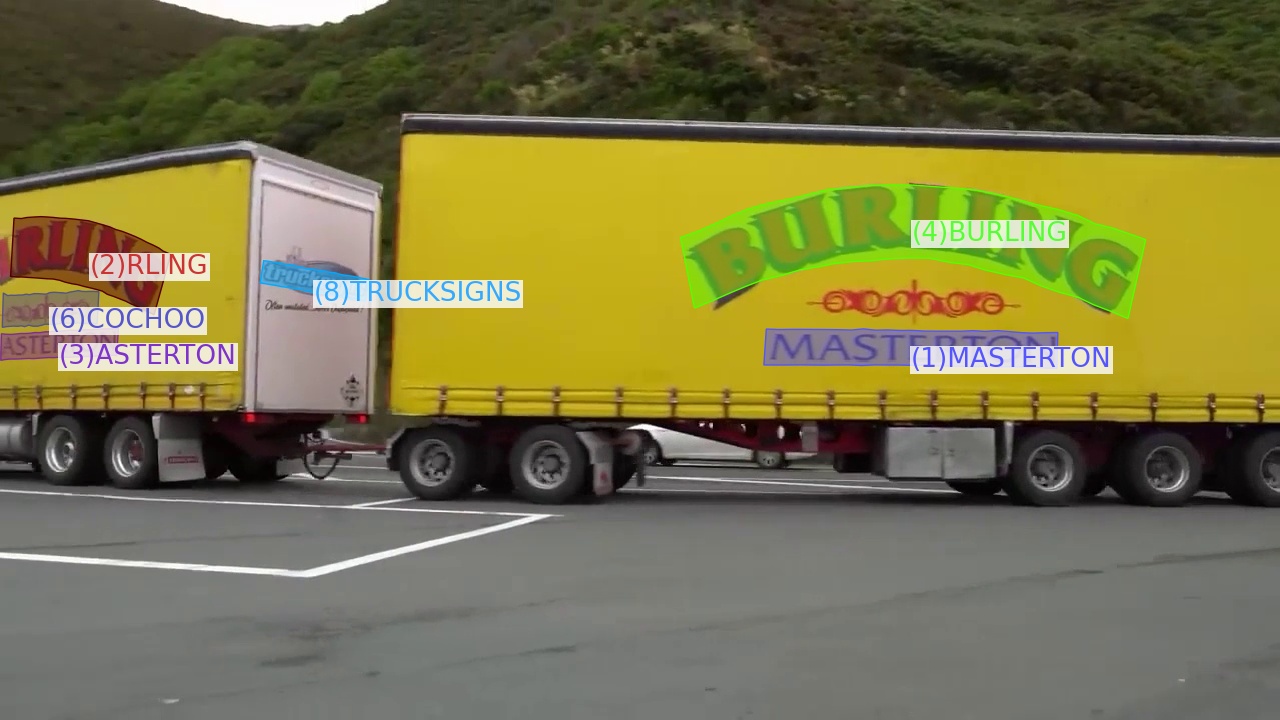}
  &\includegraphics[width=0.25\textwidth]{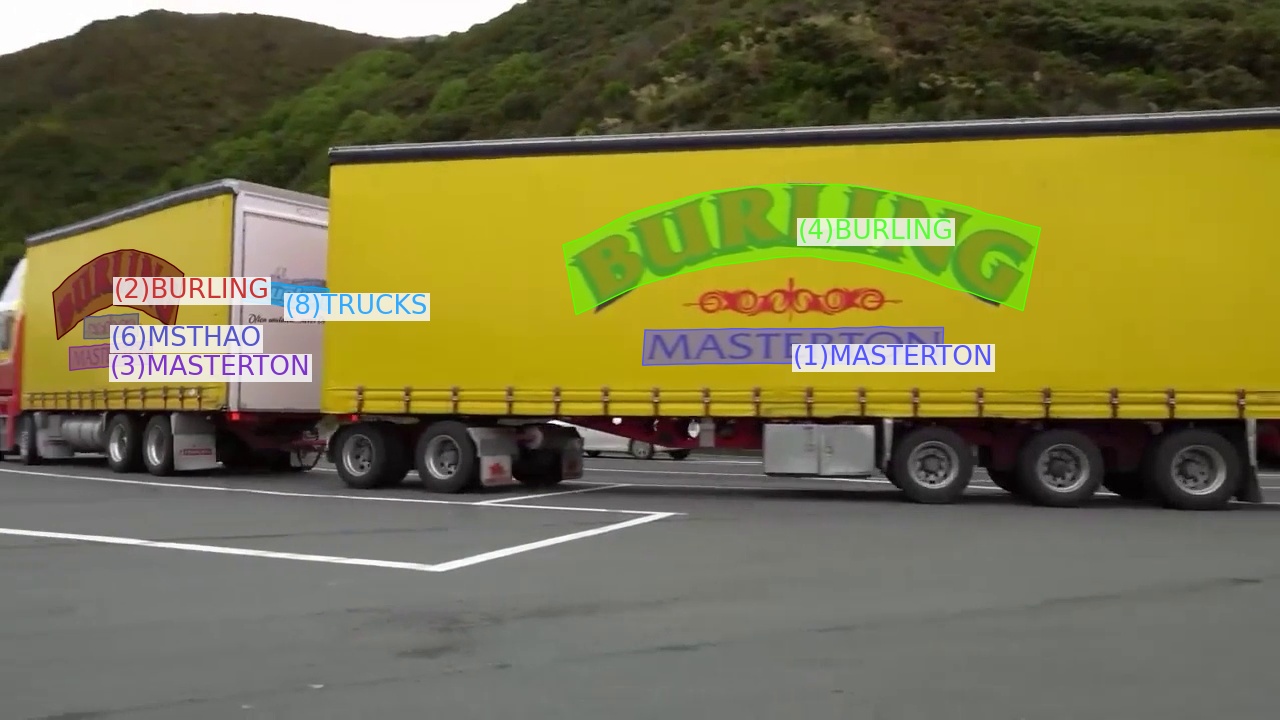}
  \\
  \end{tabular}}
\caption{\textbf{Visual results of video text spotting.} Images from top to bottom are the results on ICDAR15-video, BOVText, DSText, and ArTVideo, respectively. Text instances belonging to the same trajectory are assigned the same color.}
 \label{fig:4}
\end{figure*}

Some visual results are provided in Fig.~\ref{fig:4}. It shows that GoMatching performs well on straight and curved text, and even more complex scene text. More visual results (including some failure cases) and analysis are provided in the Appendix \ref{appx: more vis}.

\begin{table*}[t]
\centering
\caption{\textbf{Comparison results with SOTA methods on four distinct datasets.} `\dag' denotes that the results are collected from the official competition website. `*': we use the officially released model for evaluation. `M-ME' indicates whether multi-model ensembling is used. `Y' and `N' stand for yes and no. The best and second-best results are marked in \textbf{bold} and \underline{underlined}, respectively.}

\begin{minipage}{\textwidth}
\centering
\begin{minipage}{0.47\textwidth}
\centering
\subfloat[\textbf{Results on ICDAR15-video.}]{
\label{table:1}
\begin{minipage}{\linewidth}
\centering
\setlength{\tabcolsep}{2pt}
\scriptsize
\begin{tabular}{cccc}
\hline
Method &MOTA ($\uparrow$) &MOTP ($\uparrow$) &IDF1 ($\uparrow$) \\
\hline
HIK\_OCR~\cite{cheng2020free} &52.98 &74.88 &61.85 \\
CoText~\cite{Wu2022CoText} &58.94  &74.53  &71.66  \\
TransDETR~\cite{Wu2022TransDETR} &60.96 &74.61 &72.80 \\
h\&h\_lab\dag &63.76  &77.78  &71.08  \\
GOCR Offline\dag &63.05  &74.31  &76.95  \\
CoText(Kuaishou\_MMU)\dag &66.96  &76.55  &74.24  \\
\rowcolor{gray!20}
GoMatching (size:800) (ours) &68.51 &77.52  &76.59 \\
\rowcolor{gray!20}
GoMatching (size:1000) (ours) &\textbf{72.04}  &\textbf{78.53}  &\textbf{80.11}  \\
\rowcolor{gray!20}
GoMatching (size:1440) (ours) &\underline{70.52}  &\underline{78.25}  &\underline{78.70}  \\
\hline
\end{tabular}
\end{minipage}
}
\end{minipage} 
\hspace{0.02\textwidth}
\begin{minipage}{0.47\textwidth}
\centering
\vspace{-0.8cm}
\subfloat[\textbf{Results on BOVText.}]{
\label{table:3}
\begin{minipage}{\linewidth}
\setlength{\tabcolsep}{3pt}
\scriptsize
\begin{tabular}{cccc}
\hline
Method &MOTA ($\uparrow$) &MOTP ($\uparrow$) &IDF1 ($\uparrow$) \\
\hline
EAST + CRNN~\cite{Wu2021BOVText} &-79.3 &76.3 &6.8 \\
PSENet + CRNN~\cite{Wu2021BOVText} &-17.0  &79.2  &31.3  \\
DB + CRNN~\cite{Wu2021BOVText} &-13.2 &81.3 &38.8 \\
TransVTSpotter~\cite{Wu2021BOVText} &-1.4  &\underline{82.0}  &43.6  \\
CoText~\cite{Wu2022CoText} &\underline{11.4}  &80.3  &\underline{48.3}  \\
\rowcolor{gray!20}
GoMatching (ours) &\textbf{52.9}  &\textbf{87.2}  &\textbf{62.6}  \\
\hline
\end{tabular}
\end{minipage}
}
\end{minipage}
\quad
\begin{minipage}{0.47\textwidth}
\centering
\subfloat[\textbf{Results on DSText.}]{
\label{table:2}
\begin{minipage}{\linewidth}
\centering
\setlength{\tabcolsep}{1pt}
\scriptsize
\begin{tabular}{ccccc}
\hline
Method & M-ME &MOTA ($\uparrow$) &MOTP ($\uparrow$) &IDF1 ($\uparrow$) \\
\hline
TransDETR+HRNet\dag &Y &-28.58 &80.36 &26.20 \\
SCUT-MMOCR-KS\dag &Y &-27.47  &76.59  &43.61  \\
TextTrack\dag &Y &-25.09 &74.95 &26.38 \\
abcmot\dag &Y &5.54  &74.61  &24.25  \\
DA\dag &Y &10.51  &78.97  &\underline{53.45}  \\
TencentOCR\dag &Y &\underline{22.44}  &\textbf{80.82}  &\textbf{56.45}  \\
\hline
TransDETR~\cite{Wu2022TransDETR}* &N &-22.63 &79.73 &26.43 \\
\rowcolor{gray!20}
GoMatching (ours) &N &\textbf{22.83}  &\underline{80.43}  &46.09  \\
\hline
\end{tabular}
\end{minipage}
}
\end{minipage}
\hspace{0.02\textwidth}
\begin{minipage}{0.47\textwidth}
\centering
\vspace{-0.5cm}
\subfloat[\textbf{Results on ArTVideo.}]{
\label{table:4}
\begin{minipage}{\linewidth}
\setlength{\tabcolsep}{3pt}
\scriptsize
\begin{tabular}{cccc}
\hline
Method &MOTA ($\uparrow$) &MOTP ($\uparrow$) &IDF1 ($\uparrow$) \\
\hline
& \multicolumn{3}{c}{ArTVideo Tracking} \\
\hline
TransDETR~\cite{Wu2022TransDETR} &54.2 &67.9 &70.4 \\
\rowcolor{gray!20}
GoMatching (ours) &\textbf{67.2}  &\textbf{81.3} &\textbf{75.8} \\
\hline
& \multicolumn{3}{c}{ArTVideo End-to-End Spotting} \\
\hline
TransDETR~\cite{Wu2022TransDETR} &2.8 &69.7 &49.3 \\
\rowcolor{gray!20}
GoMatching (ours) &\textbf{68.8}   &\textbf{82.9}   &\textbf{78.5}  \\
\hline
& \multicolumn{3}{c}{ArTVideo-Curved Tracking} \\
\hline
TransDETR~\cite{Wu2022TransDETR} &4.4 &60.5 &50.2 \\
\rowcolor{gray!20}
GoMatching (ours) &\textbf{59.5}   &\textbf{76.3}   &\textbf{73.5}  \\
\hline
& \multicolumn{3}{c}{ArTVideo-Curved End-to-End Spotting} \\
\hline
TransDETR~\cite{Wu2022TransDETR} &-66.7 &-61.9  &26.9 \\
\rowcolor{gray!20}
GoMatching (ours) &\textbf{56.8}   &\textbf{78.0}   &\textbf{73.9}   \\
\hline
\end{tabular}
\end{minipage}
}
\end{minipage}
\end{minipage}
\label{table: comparsion results}
\end{table*}

\subsection{Ablation Studies}
\label{ablation}
We first conduct comprehensive ablation studies on ICDAR15-video to verify the effectiveness of each component. The experimental results are shown in Table~\ref{table:5}. The impact of frame length on long-term association during inference is then studied, and the results are shown in Appendix \ref{appx: frame_num}.

\noindent\textbf{Effectiveness of Utilizing Queries.} Comparing the first two rows in Table~\ref{table:5}, we can find that using queries from the decoder of image text spotter is more beneficial for tracking than RoI features. By leveraging the unified queries from frozen DeepSolo, 0.98\% and 1.05\% improvements on MOTA and IDF1 are achieved. This is because queries integrate more text instance information, \textit{i.e.}, unifying multi-scale features, text semantics, and position information, which has been proven effective in DeepSolo. Although position information is essential for tracking, it is ignored in RoI features.

\noindent\textbf{Effectiveness of Rescoring Mechanism.} To verify the effectiveness of the rescoring mechanism, we test three different scoring mechanisms: the original score from DeepSolo, the score recomputed by the rescoring head, and the fusion score from the rescoring mechanism. As shown in row 2 and row 3 of Table~\ref{table:5}, the rescoring head can alleviate the performance degradation caused by the domain gap between ICDAR15-image and ICDAR15-video, and achieve gains of 1.25\% and 0.97\% on MOTA and IDF1, respectively. 
Moreover, as shown in row 4, we can observe that combining the knowledge of rescoring head learned from the new dataset with the prior knowledge of DeepSolo can further improve MOTA and IDF1 by 0.33\% and 0.32\%, respectively.
Appendix \ref{appx: rescoring} contains more results.

\begin{figure*}[t]
\centering
  \includegraphics[width=0.9\textwidth]{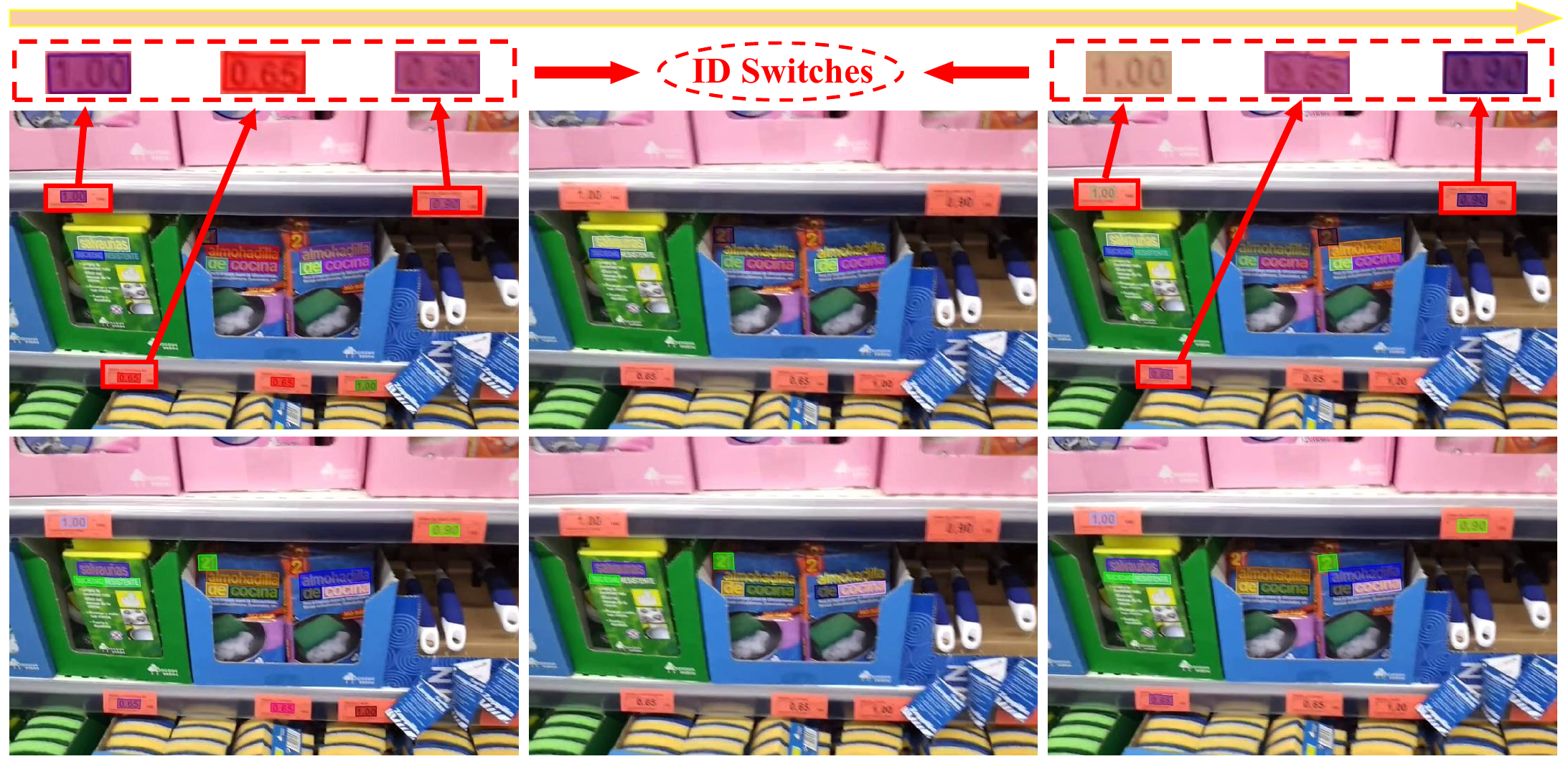}
  \caption{\textbf{Visualization results of ST-Matcher and LST-Matcher.} The first row shows the failure case that suffers from the ID switches issue when using only ST-Matcher, caused by missed detection and erroneous matching. The second row shows that LST-Matcher effectively mitigates this issue via both long and short term matching. Text instances in the same color represent the same IDs.}
 \label{fig:9} 
\end{figure*}

\noindent\textbf{Effectiveness of LST-Matcher.} In this part, we conduct three experiments to prove the effectiveness of the LST-Matcher. As shown in row 4 of Table~\ref{table:5}, we only use LT-Matcher to associate high-score text instances in the current frame with trajectories in the tracking memory bank. In row 5, we only use ST-Matcher to associate high-score text instances in the current frame with trajectories of the previous frame. In addition, as shown in row 6, we employ both LT-Matcher and ST-Matcher to test LST-Matcher.
We can easily observe that compared to LT-Matcher, LST-Matcher improves MOTA and IDF1 by 1.72\% and 1.29\% respectively, while compared to ST-Matcher, the improvement on MOTA and IDF1 are 1.12\% and 5.1\%, respectively.
In Fig.~\ref{fig:9}, we also demonstrate that using LST-Matcher can effectively mitigate the issue of ID switches caused by the strong appearance changes due to motion blur.
These results validate that combining short-term and long-term information leads to more robust tracking outcomes, thereby enhancing the performance of video text spotting.

\begin{table*}[t]
\centering
\setlength{\tabcolsep}{12pt}
\caption{Impact of difference components in the proposed GoMatching. `Query' indicates that LST-Matcher employs the queries of high-score text instances for association, otherwise RoI features. Column `Scoring' indicates the employed scoring mechanism, in which `O' means using the original scores from DeepSolo, `R' means using the scores recomputed by the rescoring head, and `F' means using the fusion scores obtained from the rescoring mechanism.
}
\resizebox{\textwidth}{!}{
\scriptsize
\centering
\begin{tabular}{c|cccc|ccc}
\hline
Index &Query &Scoring &LT-Matcher &ST-Matcher &MOTA ($\uparrow$) &MOTP ($\uparrow$) &IDF1 ($\uparrow$) \\
\hline
1 & &O &$\checkmark$ & &66.20 &78.52 &75.07 \\
2 &$\checkmark$ &O &$\checkmark$ & &67.22 &78.54 &76.12 \\
3 &$\checkmark$ &R &$\checkmark$ & &68.47 &78.29 &77.09 \\
4 &$\checkmark$ &F &$\checkmark$ & &68.80 &78.24 &77.41 \\
5 &$\checkmark$ &F & &$\checkmark$ &69.40 &\textbf{78.34} &73.60 \\
6 &$\checkmark$ &F &$\checkmark$ &$\checkmark$ &\textbf{70.52} &78.25 &\textbf{78.70} \\
\hline
\end{tabular}}
\label{table:5}
\end{table*}

\begin{table*}[t]
\centering
\setlength{\tabcolsep}{12pt}
\caption{Results of GoMatching under various training settings on the ICDAR15-video dataset. `Only Image Spotter' and `Only Tracker' refer to fine-tuning either the image spotter or tracker with another module fixed. `End-to-End' denotes that training the image spotter and tracker in an end-to-end manner. `0.001', `0.01' and `0.1' correspond to the ratios of the learning rate employed by the decoder of the image text spotter relative to the base learning rate. Due to constraints in training resources, we only optimize the parameters of the decoder component for the image text spotter.
}
\resizebox{0.95\textwidth}{!}{
\scriptsize
\centering
\begin{tabular}{c|c|ccc}
\hline
Index &Training Setting &MOTA ($\uparrow$) &MOTP ($\uparrow$) &IDF1 ($\uparrow$) \\
\hline
1 &`Only Tracker' &\textbf{72.04} &78.53 &\textbf{80.11} \\
2 &First `Only Image Spotter', Then `Only Tracker' &70.82 &78.09 &79.64 \\
3 &`End-to-End', Image Spotter's Decoder (`0.001') &71.48 & \textbf{79.14} &78.98 \\
4 &`End-to-End', Image Spotter's Decoder (`0.01') &70.15 &78.17 &77.67 \\
5 &`End-to-End', Image Spotter's Decoder (`0.1') &68.03 &75.46 &77.16 \\
\hline
\end{tabular}}
\label{table:12}
\end{table*}

\noindent\textbf{Different training strategies.} To investigate the impacts of different training strategies on GoMatching, we establish three distinct settings on the ICDAR15-video dataset, as shown in Table \ref{table:12}. 1) We only fine-tune the tracker while keeping the image text spotter frozen (the first row of Table \ref{table:12}). 2) We first fine-tune the image text spotter on images extracted from ICDAR15-video and then further fine-tune the tracker with the image text spotter fixed (the second row of Table \ref{table:12}). 3) We jointly train the spotter's decoder and tracker of GoMatching while trying different learning rates for the decoder (the last three rows of Table \ref{table:12}).
As shown in the first two rows of Table \ref{table:12}, fine-tuning the image spotter on the downstream video dataset results in a performance decline compared to the default setting. This is due to two data-related factors: \textbf{1)} minor variations in text content between frames in the same video lead to insufficient data diversity, causing the image text spotter to overfit more easily; and \textbf{2)} image blurring from camera motion reduces the quality of data available for training the image text spotter.
When the image text spotter and tracker are trained simultaneously (the last three rows), the model's performance significantly decreases. Even with the decoder's learning rate close to zero (\textit{i.e., 0.001}), there is still a 1.13\% drop in IDF1. As the decoder's learning rate increases, the performance decline becomes more pronounced. This indicates that naive joint optimization of text spotting and tracking is challenging, likely due to conflicts between the two tasks.
In future work, it is worth trying to establish larger, more diverse video text spotting datasets and to explore more effective multi-task optimization strategies.

\section{Conclusion}
In this paper, we propose a simple yet strong baseline, termed GoMatching, for video text spotting. GoMatching harnesses the talent of an off-the-shelf query-based image text spotter and only needs to tune a lightweight tracker, effectively addressing the limitations of previous SOTA methods in recognition. Specifically, we design the rescoring mechanism and LST-Matcher to adapt the image text spotter to unseen video datasets while empowering GoMatching with excellent tracking capability. Moreover, we establish a novel test set ArTVideo for the evaluation of video text spotting models on arbitrary-shaped text, filling the gap in this area. Experiments on public benchmarks and ArTVideo demonstrate the superiority of our GoMatching in terms of both spotting accuracy and training cost.

\begin{ack}
This work was supported in part by the National Key Research and Development Program of China under Grant 2023YFC2705700, in part by the National Natural Science Foundation of China under Grant U23B2048, 62076186 and 62225113, and in part by the Innovative Research Group Project of Hubei Province under Grant 2024AFA017. Dr Tao’s research is partially supported by NTU RSR and Start Up Grants. The numerical calculations in this paper have been done on the supercomputing system in the Supercomputing Center of Wuhan University.
\end{ack}


\bibliographystyle{unsrtnat}
\bibliography{ref}


\newpage

\appendix

\section{More Details of ArTVideo}
\label{appx: artvideo}
\begin{figure*}[h]
\centering
  \includegraphics[width=0.9\textwidth]{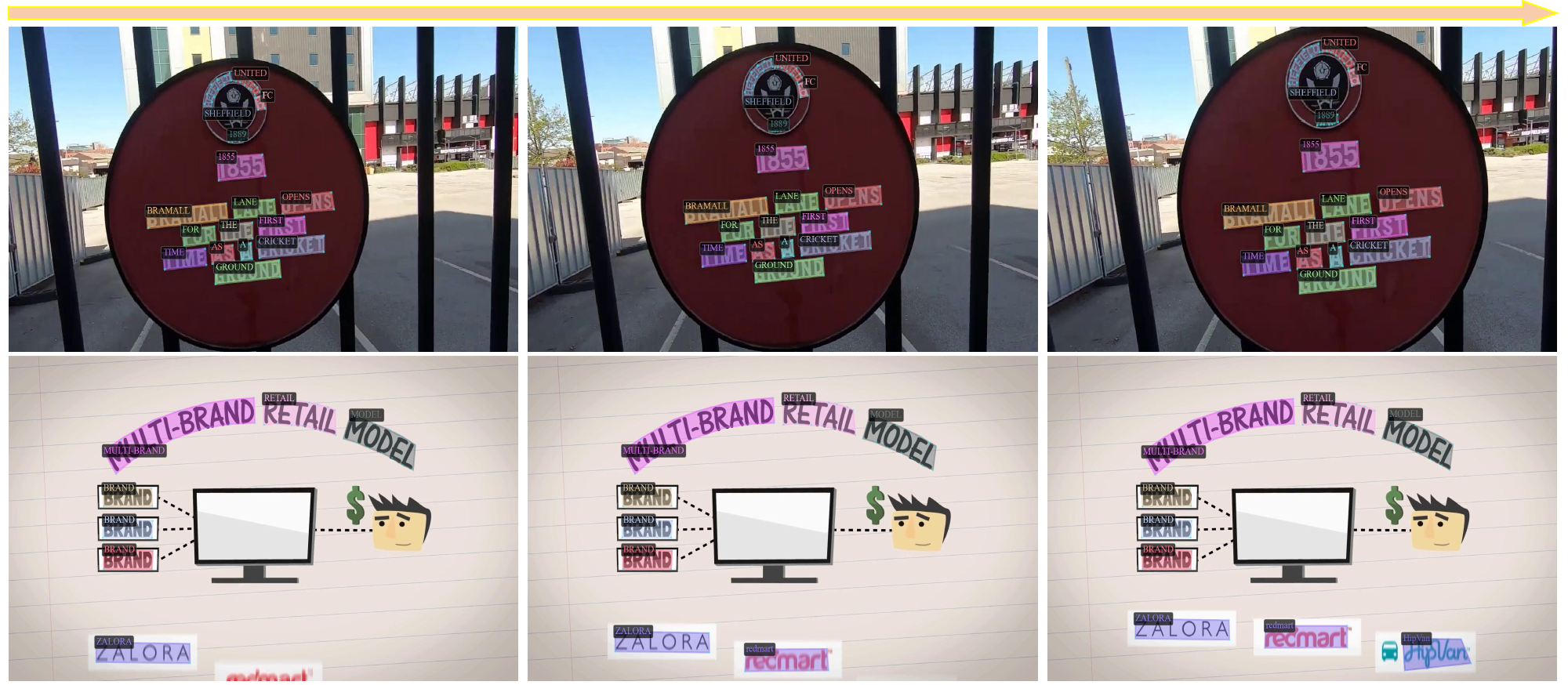}
  \caption{\textbf{Visual examples from our ArTVideo}. The straight and curved text are labeled with quadrilaterals and polygons, respectively. The same background color in different frames (columns) denotes the same instance.}
 \label{fig:5}
 \vspace{-3mm}
\end{figure*}

\begin{figure*}[h]
\centering
  \includegraphics[width=0.85\textwidth]{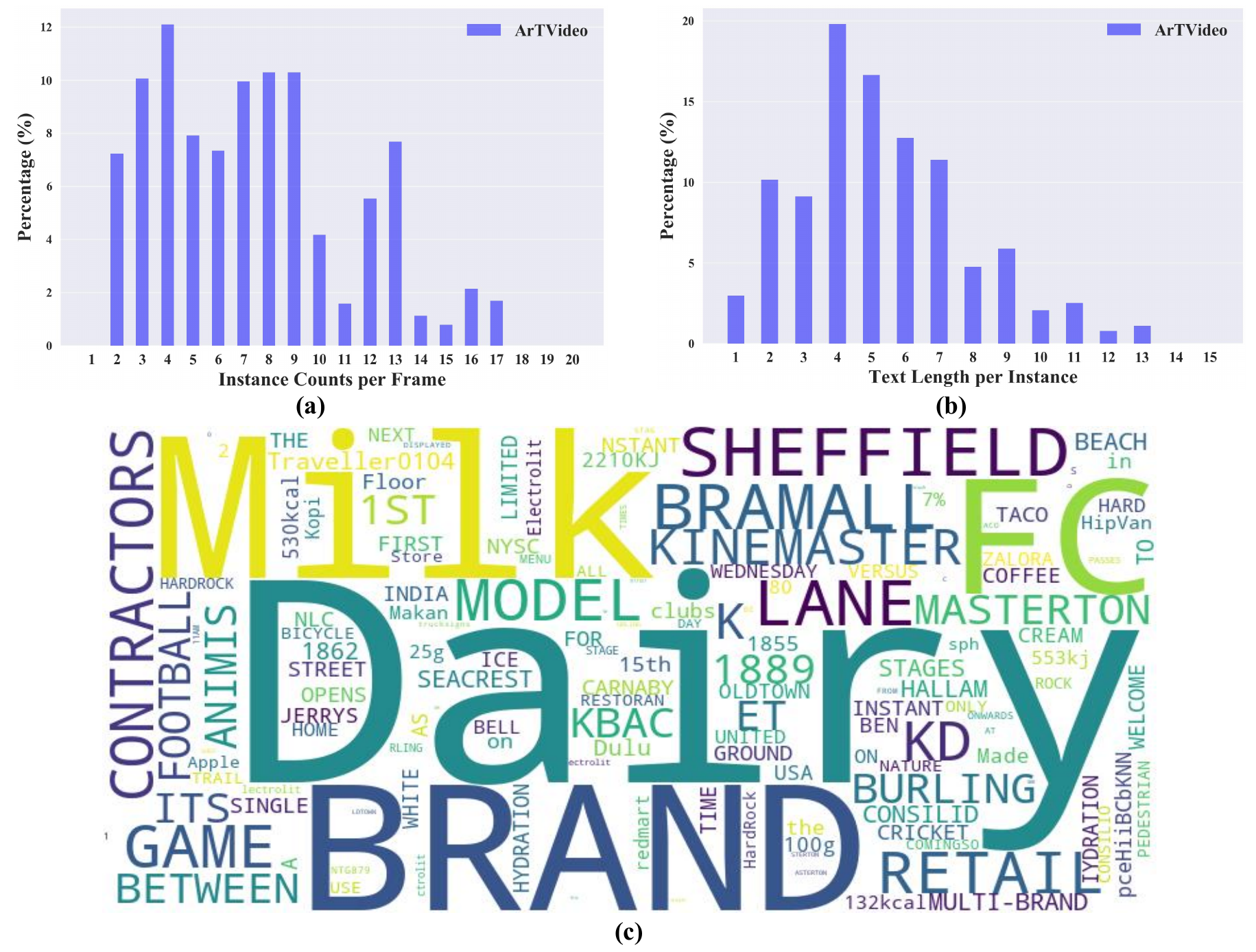}
  \caption{\textbf{Statistics of ArTVideo.} (a) and (b) show the distribution of text instance numbers in each frame and the distribution of the text length of each instance, respectively. (c) shows the word cloud of text annotations in ArTVideo.}
 \label{fig:6}
\end{figure*}

Due to the scarcity of curved text instances within existing video text spotting datasets, it is infeasible to evaluate the performance of video text spotting models on curved text. 
To fill this gap, we collected a test set named ArTVideo containing 20 video clips with a total of 884 frames, in which 18 videos were collected from YouTube and 2 videos from the BOVText test set. ArTVideo contains 6,526 text instances, including 4,632 straight text instances and 1,894 curved text instances, \textit{i.e.}, curved text accounts for about 30\%. 
As shown in Fig.~\ref{fig:5}, we provide high-quality word-level annotations for both straight and curved text in two different annotated ways. The straight text is labeled with quadrilaterals, while for curved text, we follow the CTW1500~\cite{liu2019curved} and adopt a polygon with 14 points to annotate the text contour.
More statistics about ArTVideo are shown in Fig.~\ref{fig:6}.

\section{More Details of Inference Settings}
\label{appx: settings}
Since there is no training set in ArTVideo, we directly use the model trained on ICDAR15-video to evaluate the generalization ability of GoMatching to arbitrary-shaped text. The association score threshold is set to 0.2. For ICDAR15-video, we set the shorter size of the input image to 800, 1000, and 1440, with 800 aligned with the setting in TransDETR and 1440 aligned with the setting in DeepSolo. As for BOVText, DSText, and ArTVideo, the shorter sizes are set to 1000, 1280, and 1440, respectively. All the ablation studies are conducted on the setting of 1440.

\section{Impact of the Frame Number in LT-Matcher}
\label{appx: frame_num}
In Table~\ref{table:6}, we further study and analyze the impact of the number of frames for long-term association in LT-Matcher during inference. For the text spotting task, since a single frame may have a large number of text instances, sometimes reaching hundreds, excessive historical frame information would weaken the discrimination of text instance features, resulting in erroneous matching results. Therefore, we conduct a hyper-parameter search and find that the optimal frame number is 6.

\begin{table*}[t]
\begin{minipage}{0.5\textwidth}
    \setlength{\tabcolsep}{5pt}
    \footnotesize 
    \centering
    \caption{Ablation studies on the number of frames ($T$) for long-term association in LT-Matcher, and the max number of history frames in tracking memory bank is $H=T-1$). Experiments are conducted on ICDAR15-video and the best results are marked in \textbf{bold}.}
    \begin{tabular}{cccc}
    \hline
    Number $T$ &MOTA ($\uparrow$) &MOTP ($\uparrow$) &IDF1 ($\uparrow$) \\
    \hline
    $T = 32$ &70.13 &78.07 &78.24 \\
    $T = 16$ &70.33 &78.25  &78.60  \\
    $T = 8$ &70.44 &78.25 &\textbf{78.70} \\
    $T = 6$ &\textbf{70.52}  &78.25  &\textbf{78.70}  \\
    $T = 4$ &70.51  &\textbf{78.27}  &78.16  \\
    \hline
    \end{tabular}
    \label{table:6}
\end{minipage}
\hfill
\begin{minipage}{0.47\textwidth}
    \setlength{\tabcolsep}{5pt}
    \footnotesize 
    \centering
    \caption{\textbf{Results of different score fusion strategies on ICDAR5-video.} `Mean', `Geo-mean', and `Maximum' denote the arithmetic mean, geometric mean, and the maximum score fusion strategies, respectively. The best results are highlighted in \textbf{bold}.}
    \begin{tabular}{cccc}
    \hline
    Strategy &MOTA ($\uparrow$) &MOTP ($\uparrow$) &IDF1 ($\uparrow$) \\
    \hline
    Mean &70.46  &78.38  &78.29  \\
    Geo-mean &70.29  &\textbf{78.39}  &78.26  \\
    Maximum &\textbf{70.52}  &78.25  &\textbf{78.70}  \\
    \hline
    \end{tabular}
    \label{table:7}
\end{minipage}
\end{table*}

\begin{table*}[t]
\begin{minipage}{0.5\textwidth}
    \setlength{\tabcolsep}{2pt}
    \scriptsize 
    \centering
    \caption{\textbf{Results of using different image sizes on ICDAR15-video.} `Size' means the size of the shorter side of the input image during inference. The best results are highlighted in \textbf{bold}.}
    \begin{tabular}{ccccc}
    \hline
    Method &MOTA ($\uparrow$) &MOTP ($\uparrow$) &IDF1 ($\uparrow$) &FPS ($\uparrow$)\\
    \hline
    TransDETR(Size: 800) &60.96 &74.61 &72.80 &12.69  \\
    GoMatching(Size: 800) &68.51  &77.52  &76.59 &\textbf{14.41} \\
    GoMatching(Size: 1000) &\textbf{72.04}  &\textbf{78.53}  &\textbf{80.11} &10.60 \\
    \hline
    \end{tabular}
    \label{table:8}
\end{minipage}
\hfill
\begin{minipage}{0.47\textwidth}
    \setlength{\tabcolsep}{6pt}
    \footnotesize 
    \centering
    \caption{\textbf{Comparison between TransDETR and GoMatching.} `T-Para.' and `A-Para.' denote the number of all parameters and the trainable parameters in each model, respectively.}
    \begin{tabular}{cccc}
    \hline
    Method & \#T-Para. (M) & \#A-Para. (M) \\
    \hline
    TransDETR &39.35  &39.58  \\
    GoMatching &32.79  &75.38  \\
    \hline
    \end{tabular}
    \label{table:9}
\end{minipage}
\end{table*}

\begin{figure*}[t]
    \centering
    \includegraphics[width=0.9\textwidth]{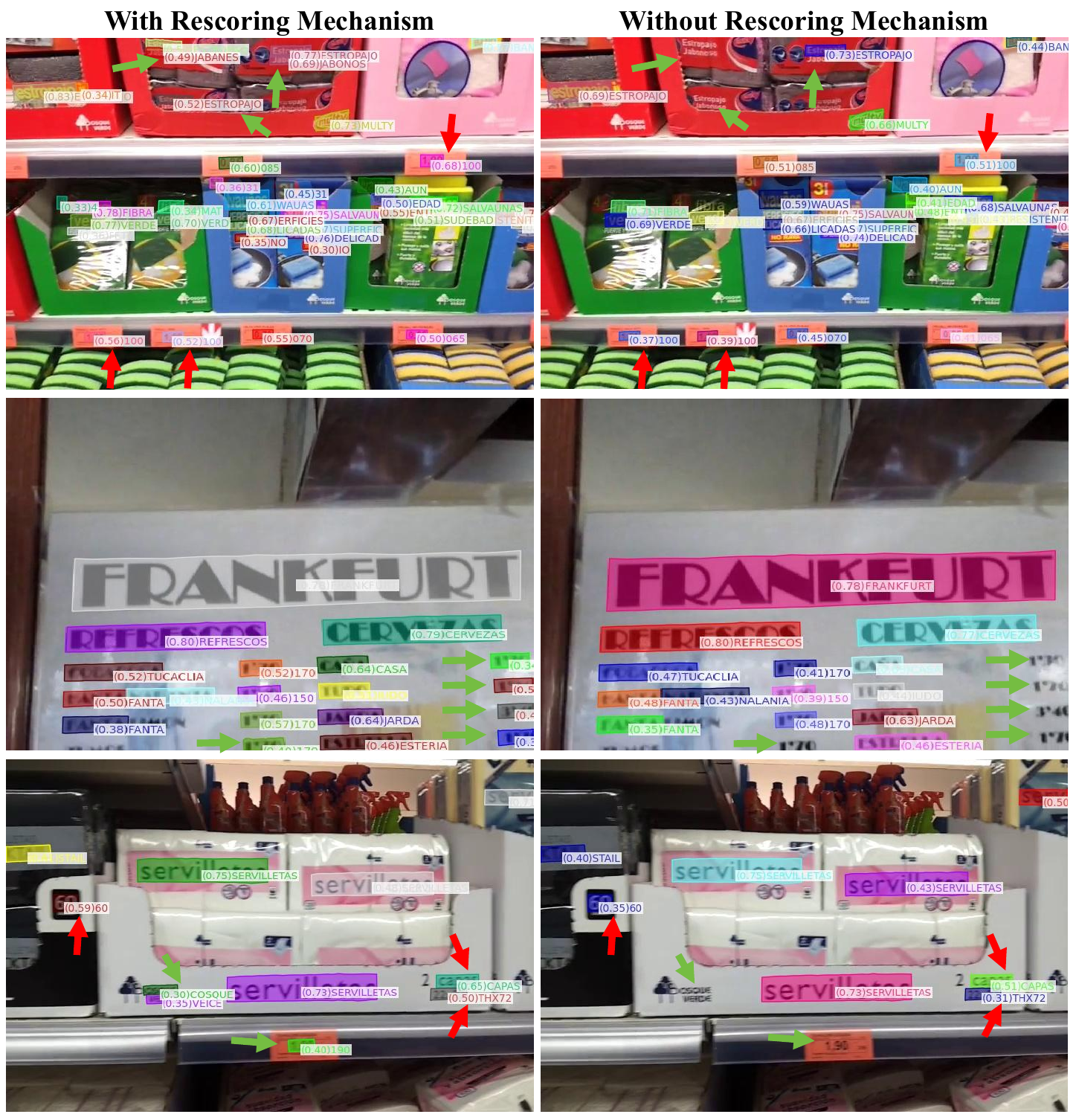}
    \caption{\textbf{Visual comparison of GoMatching with and without the rescoring mechanism.} The values in parentheses indicate the confidence scores of the detected text. `{\color{green}$\rightarrow$}' points to the filtered texts due to low confidence without using rescoring. `{\color{red}$\rightarrow$}' points to the text whose confidence score has been improved by the rescoring mechanism. The rescoring mechanism increases the confidence scores of small texts and blurry texts, preventing them from being filtered out by the threshold and thereby cultivating a better tracking candidate pool.}
    \label{fig:7}
\end{figure*}

\begin{table*}[t]
\begin{minipage}{0.5\textwidth}
    \setlength{\tabcolsep}{2pt}
    \scriptsize 
    \centering
    \caption{\textbf{Video text detection performance on ICDAR2013-video ~\cite{karatzas2013icdar}.} The best results are highlighted in \textbf{bold}.}
    \begin{tabular}{cccc}
    \hline
    Method &Precision ($\uparrow$) &Recall ($\uparrow$) &F-measure ($\uparrow$) \\
    \hline
    Free &79.7 &68.4 &73.6 \\
    TransDETR &80.6  &70.2  &75.0 \\
    GoMatching w/o rescoring &\textbf{92.4}  &65.7  &76.8 \\
    GoMatching &89.5  &\textbf{74.8}  &\textbf{81.5} \\
    \hline
    \end{tabular}
    \label{table:10}
\end{minipage}
\hfill
\begin{minipage}{0.47\textwidth}
    \setlength{\tabcolsep}{2pt}
    \scriptsize 
    \centering
    \caption{\textbf{Comparison results of detection AP on the ICDAR13-video between with and without the rescoring mechanism.}}
    \begin{tabular}{ccccc}
    \hline
    Method &AP &$\text{AP}_\text{S}$ &$\text{AP}_\text{M}$ &$\text{AP}_\text{L}$\\
    \hline
    w/o rescoring &26.2  &11.6  &40.1  &49.8  \\
    w/ rescoring &29.3 (\darkgreen{+3.1})  &15.5 (\darkgreen{+3.9})  &42.7 (\darkgreen{+2.6})  &51.9 (\darkgreen{+2.1})  \\
    \hline
    \end{tabular}
    \label{table:11}
\end{minipage}
\end{table*}

\section{Comparison of Different Score Fusion Strategies}
To investigate the impact of the score fusion strategy in the rescoring mechanism, we further evaluate two other strategies: (1) the arithmetic mean score fusion strategy and (2) the geometric mean score fusion strategy, denoted as mean and geo-mean, respectively. The arithmetic mean score fusion strategy takes the average of the scores from the image text spotter and the rescoring head as the final score for each query, while the geo-mean score fusion strategy uses the geometric mean. These two strategies can be formulated as:
\begin{equation}
    c_{mean} = (c_o + c_r) / 2,
    \label{eq:13}
\end{equation}
\begin{equation}
    c_{geo-mean} = \sqrt{c_o * c_r},
    \label{eq:14}
\end{equation}
where $c_{mean}$ and $c_{geo-mean}$ denote the final score of the two strategies, respectively. $c_o$ and $c_r$ are the scores from the image text spotter and the rescoring head, respectively.

From Table~\ref{table:7}, we can see that employing the maximum score fusion strategy achieves the best performance on MOTA and IDF1 among all the strategies. As for the other two strategies, extremely low confidence scores from the image text spotter may result in low final scores, probably leading to missed detections. Therefore, we adopt the maximum score fusion strategy in the rescoring mechanism of GoMatching by default.

\section{More Comparisons between TransDETR and GoMatching}
In Table~\ref{table:8}, we provide the results of using three different image sizes in GoMatching during inference on ICDAR15-video. In the first row of Table~\ref{table:8}, the shorter side of the input image is set to 800, which is the same as the default setting of TransDETR. It is evident from Table~\ref{table:8} that GoMatching significantly outperforms TransDETR in all settings. Meanwhile, it also shows that GoMatching outperforms TransDETR under its default setting in terms of both inference speed and spotting accuracy. With the increase in image size (\textit{e.g.}, 1000), GoMatching offers better spotting performance at the cost of decreased inference speed.  

Moreover, we compare the number of all parameters and the trainable parameters of GoMatching and TransDETR, as shown in Table~\ref{table:9}. 
It is noteworthy that GoMatching requires a much smaller training budget than TransDETR owing to its simpler architecture design, making it a simple but strong baseline for video text spotting.

\begin{figure*}[t]
\centering
  \includegraphics[width=0.9\textwidth]{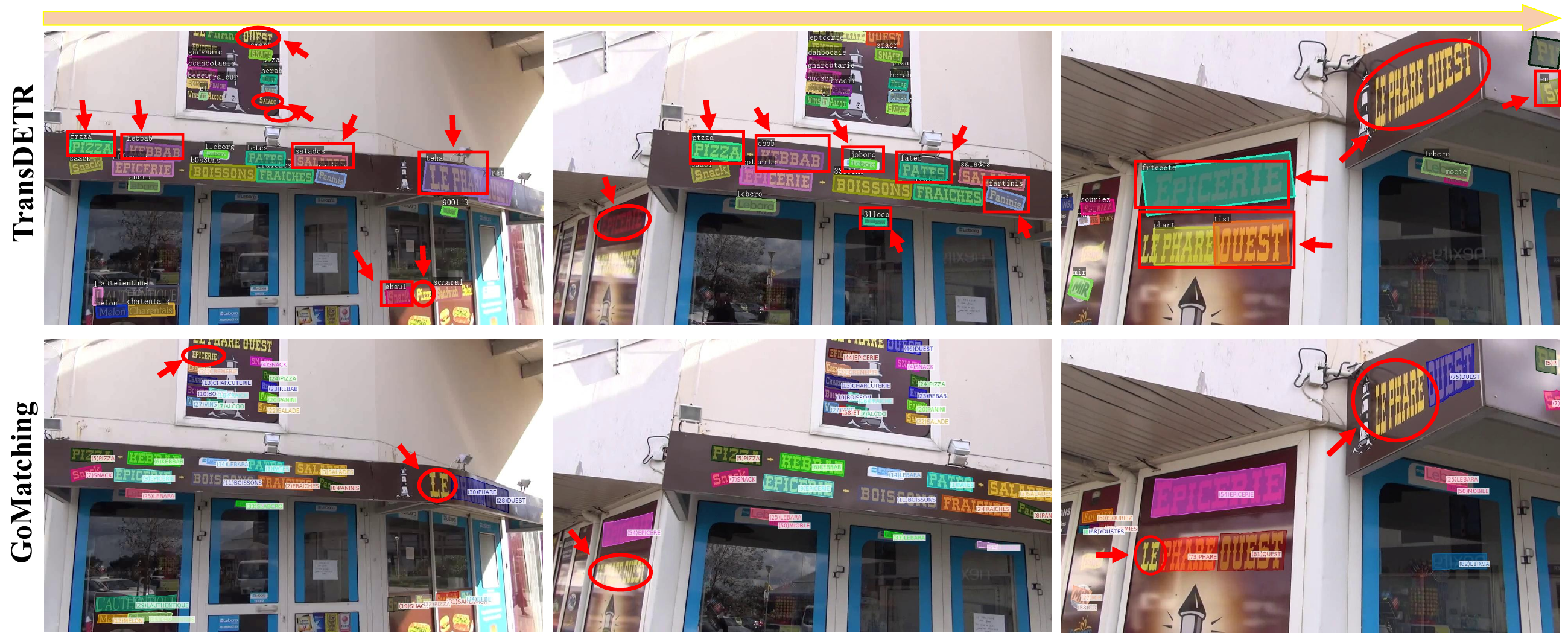}
  \caption{\textbf{More visualization results of TransDETR and GoMatching.} Failure cases of text detection and recognition are highlighted with ellipses and rectangles, respectively.}
 \label{fig:8}
\end{figure*}

\section{More Results for the Rescoring Mechanism}
\label{appx: rescoring}
In Table ~\ref{table:10}, we provide the video text detection results of GoMatching on ICDAR13-video ~\cite{karatzas2013icdar} and compare them with Free and TransDETR. As shown in the table, without the rescoring mechanism, GoMatching relies on the original results from DeepSolo, resulting in a 4.5\% decrease on Recall and only a 1.8\% improvement on F-measure compare to TransDETR. This is due to the domain gap between image data and video data, directly using an image spotter leads to a low confidence and consequently low Recall on video data. When encompasses the rescoring mechanism, GoMatching achieves a 9.1 improvement on Recall compared to DeepSolo and a 6.5\% F-measure enhancement compared to TransDETR. These improvements highlight the effectiveness of rescoring mechanism in alleviating the domain gap and leading to a better tracking candidate pool. The impressive results of GoMatching are not merely attributed to the introduction of a robust image text spotter.

To further explore how the rescoring mechanism eases the domain gap, we calculate the AP of detection results on ICDAR13-video, as shown in the Table ~\ref{table:11}. Incorporating the rescoring mechanism effectively improves the detection performance of DeepSolo on video datasets, particularly for small text, resulting in a 3.9\% improvement on $\text{AP}_\text{S}$. We also present more visual results to embody the potency of rescoring mechanism in Fig.~\ref{fig:7}.

\section{More Visualization Results}
\label{appx: more vis}
We present more visualization results of TransDETR and GoMatching in Fig.~\ref{fig:8}, including some failure cases. It can be observed that GoMatching exhibits a significant improvement in text recognition performance compared to TransDETR.
It should be noted that GoMatching may also experience failures due to the image-video domain gap and extreme cases, such as very small text instances and significant motion blur. These issues can be mitigated by employing a stronger text spotter with a more representative backbone and training on larger, more diverse datasets.

\end{document}